\documentclass[10pt]{article}

\usepackage[preprint]{tmlr}
\usepackage[utf8]{inputenc}
\usepackage{microtype}
\usepackage{inconsolata}

\usepackage{hyperref}
\usepackage{url}
\usepackage{booktabs}
\usepackage{amsfonts}
\usepackage{amsmath}
\usepackage{amssymb}
\usepackage{amsthm}
\usepackage{nicefrac}
\usepackage{float}
\usepackage{needspace}

\usepackage{graphicx}
\graphicspath{{figures/}{./}}
\usepackage{xcolor}
\definecolor{tocblue}{RGB}{40,70,160}
\usepackage[most]{tcolorbox}
\definecolor{boxframe}{RGB}{170,170,170}
\definecolor{boxtitlebg}{RGB}{208,208,208}
\tcbset{graybox/.style={
  breakable, enhanced, colback=gray!3, colframe=boxframe,
  colbacktitle=boxtitlebg, coltitle=black, fonttitle=\bfseries\small,
  arc=0.6mm, boxrule=0.4pt, left=2mm, right=2mm, top=1mm, bottom=1mm}}
\newtcolorbox{promptbox}[1][]{graybox, fontupper=\footnotesize\ttfamily, title={#1}}
\usepackage{multirow}
\usepackage{xspace}
\usepackage{tikz}
\usetikzlibrary{positioning, arrows.meta, calc}

\newtheorem{definition}{Definition}

\newcommand{\rr}{\textsc{RR}\xspace}
\newcommand{\lossy}{\textsf{lossy}\xspace}
\newcommand{\padded}{\textsf{lossy-padded}\xspace}
\newcommand{\srcfirst}{\textsf{source-first}\xspace}

\makeatletter
\newcommand{\reporef}{\if@preprint\footnote{Code, data, and the reproduction harness: \url{https://github.com/collapseindex/reclaim-eval}.}\else\footnote{Code, data, and the reproduction harness are released; repository link withheld for double-blind review.}\fi}
\makeatother

\title{Reclaim Evaluation: A Lossy Memory Is Worse Than an Empty One}

\author{\name Alex Kwon \email ask@collapseindex.org \\
\addr Independent Researcher}

\begin{document}

\maketitle

\begin{abstract}
A language model's memory can be \textbf{worse than no memory at all} when the model or its interface is
disposed to act on it: a memory that keeps a wrong conclusion but drops the work behind it leads a model
to re-emit the stale value as a confident answer, where an empty memory leads it to abstain. We call this
\textbf{brittle memory}. The information loss is definitional; the finding is behavioral, and it turns on
one thing, whether the memory kept a \emph{re-derivation basis} (the source) rather than the answer. We
measure it with \textbf{reclaim evaluation}: induce a known drift, compress at a fixed budget, deliver a
correction that names the error, and score exact recovery, judge-free. Holding the budget fixed and
varying only \emph{what} the compression keeps isolates correctability from capability and from size; an
$8$B model and a frontier one wall in the same place. A one-line \textbf{source-first} policy, keep the
recomputable source, drop the re-derivable conclusion, restores correctability at equal budget where the
source is compact and identifiable, with a length-matched control ruling out ``more text.'' We map where
the fix fails, show the failure compounds through memory loops, and replicate across three deployed memory
systems, real dialogue (MultiWOZ), and $\tau$-bench, a deployed-agent benchmark where whether a lossy
memory becomes a \emph{harmful action} is a joint model-and-interface property. We release the harness,
the paired memory conditions, and validators built to come out false.\reporef\footnote{AI assistants were used for coding support and for drafting and editing
manuscript prose; all research questions, experimental design, analyses, and conclusions are the
authors' own. See the \emph{Use of AI Assistants} statement for details.}
\end{abstract}

\begin{figure}[t]
\centering
\includegraphics[width=\textwidth]{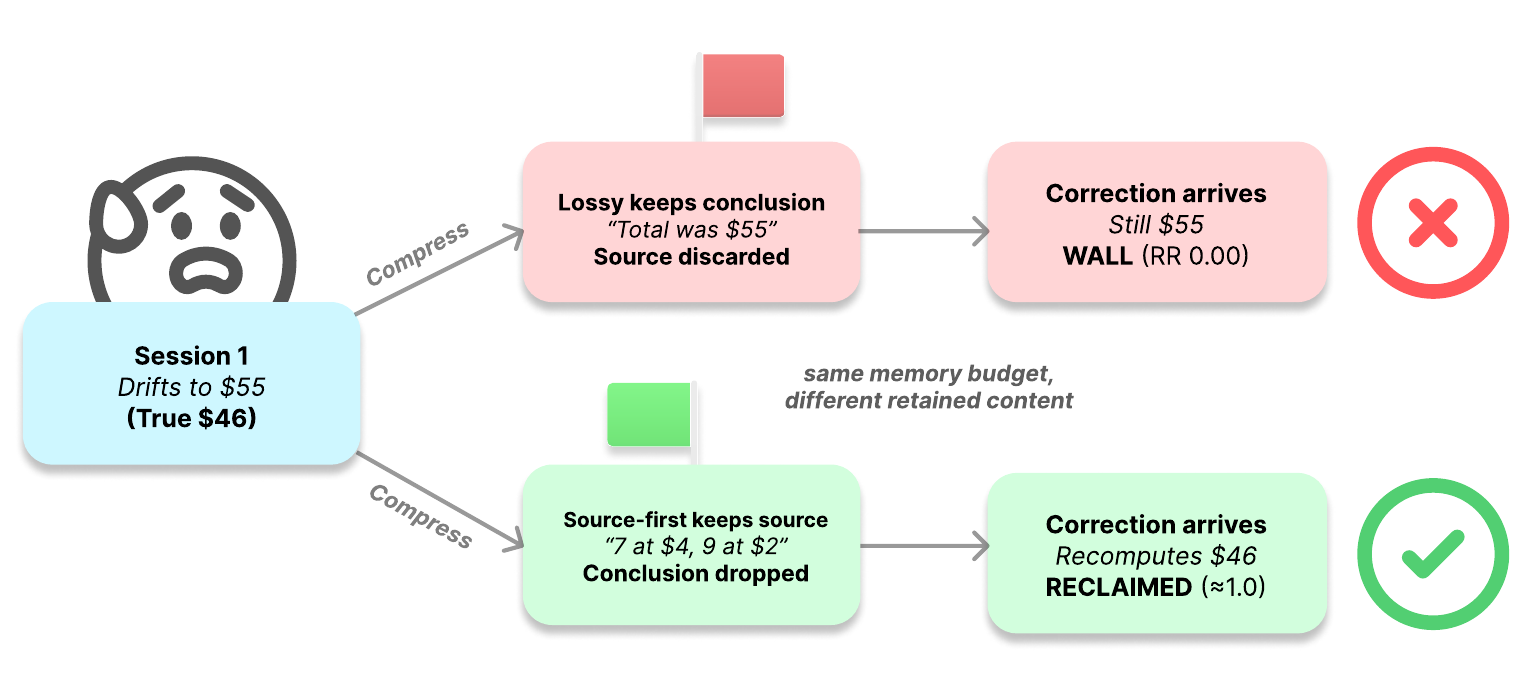}
\caption{\textbf{Compression decides whether an error stays fixable.} Only a compressed memory crosses
from session~1 to session~2, at a fixed budget. \textbf{\lossy{}} compression keeps the salient
\emph{wrong} conclusion and discards the source, so a later correction has nothing to recompute from and
the model confidently returns the stale value; \textbf{\srcfirst{}} keeps the recomputable source instead,
so the same correction lands and the model recovers the truth. Same budget, opposite outcome. Numbers are
directed-arm \rr{} at low integrity; the \srcfirst{} memory shown is the hand-built oracle, and a
deployable one-prompt distiller reclaims $0.49$--$0.88$ with a weak writer, rising to $1.00$ when a
frontier model writes it (\S\ref{sec:generality}).}
\label{fig:teaser}
\end{figure}

\section{Introduction}
\label{sec:intro}

Language models are increasingly deployed in settings where they must remember. Assistant
``memory'' features, long-running agents, and retrieval pipelines all carry information
forward across a context window or a session boundary, and they do so by
\emph{compressing}: a conversation, a document, or a trajectory is reduced to a summary, a
note, or a set of retrieved chunks \citep{packer2024memgpt, lewis2020retrieval}. The
implicit assumption behind these systems is that a compression which preserves the model's
answer has preserved what matters.

We show that this assumption is incomplete in a consequential way. Compression decides not
only what a model can recall, but whether the model can later be \emph{corrected}. When a
model has committed to a wrong intermediate conclusion and that conclusion is carried
forward while the evidence behind it is dropped, a later correction has nothing to act on. The model
restates the error, and no stronger model we tested recovers. The reason is structural: a correction acts
on the \emph{computation} that produced the answer, not on the answer itself. A memory therefore stays
correctable only if it kept a \emph{re-derivation basis} (the source); the answer alone is a fixed point
of correction. This one principle predicts the results that follow, including a non-obvious
\emph{capability inverted-U}: a stronger reader helps most only where the memory kept a partial
recomputation path (\S\ref{sec:principle}). We call the failure \textbf{brittle memory}: the salient
answer is still there, yet it shatters the moment it must support a correction.

The failure also compounds, because agents feed memory into memory. A single dropped-source error corrupts
a growing span of downstream steps, and no correction that names the error, however late it arrives, 
repairs them, while a source-preserving memory stays correctable to a budget horizon
(\S\ref{sec:cascade}). The stake is thus not one wrong answer but a compounding, uncorrectable one.

This is not hypothetical: the dominant instinct in summarization is to keep the
\emph{takeaway} and drop the \emph{working}. A note that records ``the total was \$55''
while discarding the line items is precisely the failure above, and we show it is a design
choice rather than a property of the model.

Two things must be kept apart. The information loss is immediate from the definition, a value cannot
be recomputed without its inputs, and no capability changes that. That is only the \emph{setup}. The
\emph{finding} is behavioral: the same model abstains or emits a confident wrong value depending only on
what the memory kept. Calibration here is induced by a memory design, not a fixed trait to measure once
(\S\ref{sec:results}). We do not propose universal source retention. We identify the regime where memory
must keep a recomputation path, and show that summary-like memories often drop it exactly there; how much
real assistant memory lives in that regime scales the failure's reach in practice, a question our audit
orders but does not size (\S\ref{sec:limitations}).

To study correctability directly we introduce \textbf{reclaim evaluation}
(\S\ref{sec:method}). We induce drift on a task with a known answer, deepen the model's
commitment over several turns, compress the interaction into a single carried memory at a
fixed budget, and then deliver a \emph{directed} correction that names the error
without supplying the answer. The measured quantity is the Reclaim Rate (\rr): how often
the correction recovers the truth. By holding the budget fixed and varying only \emph{what}
the compression keeps, we isolate correctability from both model capability and memory
size. Table~\ref{tab:claims} maps every load-bearing claim to its evidence and epistemic status
(\textsc{shown}, \textsc{analytic}, or \textsc{suggestive}).

\clearpage
\begin{table}[H]
\centering\small
\caption{\textbf{Claims and evidence.} Every load-bearing claim, the evidence behind it, and its
epistemic status: \textsc{shown} (direct measurement), \textsc{analytic} (follows from the definitions),
or \textsc{suggestive} (preliminary, not load-bearing). The behavioral results are the contribution; the
information bound (\textsc{analytic}) is apparatus.}
\label{tab:claims}
\begin{tabular}{@{}p{0.66\textwidth} l l@{}}
\toprule
\textbf{Claim} & \textbf{Evidence} & \textbf{Status} \\
\midrule
A lossy memory is never better than an empty one, and strictly worse on models disposed to answer: with the source gone, a model that answers emits a confident wrong value instead of abstaining (4 of 8 models; the OpenAI and Anthropic systems, incl.\ the small \texttt{gpt-4o-mini}, abstain and escape it; the split tracks vendor/training family rather than scale, in the snapshots evaluated). & Tab.\ \ref{tab:blank},\,\ref{tab:attractor},\,\ref{tab:failmode},\,\ref{tab:disposition} & \textsc{shown} \\
The frontier ``escape'' is an interface affordance: through a mandatory structured-output field (no abstain token, as production tool calls have), the abstaining models commit the inherited wrong value $0.72$--$1.00$ of the time under lossy and $0.00$ under empty. & Tab.\ \ref{tab:interface} & \textsc{shown} \\
The wall and fix sit in the same place from an $8$B model to a frontier one (capability-invariant). & Tab.\ \ref{tab:wall},\,\ref{tab:frontier} & \textsc{shown} \\
Capability's benefit is non-monotone in how much source survives (the inverted-U): nil at the wall, peaking at a partial-source wall, swept within arithmetic by degrading the source at fixed budget (gap $+0.46$ at partial vs $0.00$ at both ends). & Fig.\ \ref{fig:invertedu}, Tab.\ \ref{tab:frontier} & \textsc{shown} \\
The same non-monotonicity turns mildly negative in an agentic task ($\pi{=}0$), though capability there is confounded with decoding determinism. & Tab.\ \ref{tab:battleship} & \textsc{suggestive} \\
Source-first removes the wall at equal budget; the lever is content, not text. & Tab.\ \ref{tab:wall} & \textsc{shown} \\
A single \lossy{} error cascades across a memory loop and stays uncorrectable short of re-supplying the answer; source-first holds to a budget horizon. & Tab.\ \ref{tab:cascade} & \textsc{shown} \\
The wall and fix appear in three deployed memory systems and on real dialogue (MultiWOZ). & Tab.\ \ref{tab:frontier},\,\ref{tab:multiwoz} & \textsc{shown} \\
The wall and source-first fix replicate on a real deployed-agent benchmark ($\tau$-bench), judge-free (database-state scoring) and scale-invariant from a $3$B model to the frontier (\lossy{} \rr{} $\leq 0.14$ at every scale; \srcfirst{} recovers the action $0.60$--$1.00$, tracking reader capability). & Tab.\ \ref{tab:taubench} & \textsc{shown} \\
Under a mandatory tool interface, whether a \lossy{} memory becomes a \emph{harmful action} is a model-by-interface property: given a safe-exit tool the frontier escalates (harm $\leq 0.07$) while open models and \texttt{gpt-4o-mini} commit the wrong action ($0.78$--$0.96$, a $70$B model included, so not scale); remove the safe option and three of four frontier models commit the error too, only \texttt{claude-opus-4.8} refusing (an intrinsic, recoverability-aware refusal). & Tab.\ \ref{tab:forced} & \textsc{shown} \\
The wall reproduces in an agentic task: without its own shot record a model refires $0.77$--$0.91$ of the time and never abstains (a silent action), collapsing its hit rate. & Tab.\ \ref{tab:battleship} & \textsc{shown} \\
A one-prompt source-first deploys ($0.49$--$0.88$); aiming retrieval at the source instead does not rescue it (distillation is the lever). & Tab.\ \ref{tab:frontier} & \textsc{shown} \\
Source-first is robust to generic, false-locus, and confident-wrong-value corrections; a sustained push and an injected fabricated source are capability-gated. & Tab.\ \ref{tab:generic},\,\ref{tab:adversarial} & \textsc{shown} \\
Past its budget or under noise the fix decays (the crossover tracks the budget) and fails silently unless the note records completeness. & Fig.\ \ref{fig:sizesweep},\,\ref{fig:noisy}; Tab.\ \ref{tab:complete} & \textsc{shown} \\
A write-time recompute certificate flags the silent size-truncation failure capability-free (reclaim $0.93$ passing vs $0.00$ flagged), a near-complete guard with a small noise-boundary residual. & \S\ref{sec:certificate} & \textsc{shown} \\
Correctability tracks retention of a re-derivation basis, not capability or budget: the answer is a fixed point of correction, the source is not. & Def.\ (\S\ref{sec:principle}) & \textsc{analytic} \\
\emph{Apparatus}: source absent $\Rightarrow$ the answer is uncomputable; the wall exists. & Def.\ \& validators (\S\ref{sec:method},\,\ref{sec:setup}) & \textsc{analytic} \\
Compact-source memory is more prevalent in agentic than chat domains (robust ordering under two LLM labelers and a judge-free deterministic floor of $0.86$ on agentic; absolute level not identified). & Tab.\ \ref{tab:prevalence} & \textsc{suggestive} \\
\bottomrule
\end{tabular}
\end{table}
\clearpage

\needspace{5\baselineskip}
\paragraph{Contributions.}
\begin{itemize}
\itemsep0.3em
\item \textbf{A predictive account} (\S\ref{sec:principle}). Correction acts on the computation, so a
memory stays correctable only if it keeps a \emph{re-derivation basis}, not the answer. The principle
predicts rather than restates the results: the wall, its capability-invariance, the length control, and
the boundaries all follow from how much of the basis a memory keeps.
\item \textbf{The capability inverted-U.} The principle predicts a non-obvious regularity: a stronger
reader's benefit is \emph{non-monotone} in how much of the basis survives, nil at the wall and greatest
on a \emph{partial}-source wall. We confirm it as a within-task swept curve ($+0.46$ at partial, $0.00$
at both ends; Fig.~\ref{fig:invertedu}) and across deployed memory systems (\S\ref{sec:generality}).
\item \textbf{The finding, and the method that isolates it.} \emph{Reclaim evaluation} compresses a
committed interaction under matched-budget policies that differ only in what they retain, separating
correctability from capability and budget (\S\ref{sec:method}). With it: a lossy memory is worse than an
empty one, because a model disposed to answer emits the stale value rather than abstaining (the
\textbf{brittle memory} regime). The wall is capability-invariant (an $8$B and a frontier model fail in
the same place); the emit-versus-abstain response is set by disposition, not capability (eight models,
App.~\ref{app:disposition}).
\item \textbf{The fix, and its regime.} A one-line \emph{source-first} policy (keep the recomputable
source, drop the re-derivable conclusion) removes the failure at equal budget where the source is compact
and identifiable; a length-matched control rules out ``simply more text,'' and the deployable one-prompt
form is materially weaker than the oracle ($0.49$--$0.88$). Two sweeps locate where it decays to the lossy
floor (source size, decoy noise), and past that boundary it fails \emph{silently} unless a one-line
completeness signal restores loud failure (\S\ref{sec:boundary}).
\item \textbf{The deployment stake, and replication.} When memory feeds memory, a single \lossy{} error
corrupts a growing blast radius that resists any locus-naming correction, while source-first holds to a
budget horizon (\S\ref{sec:cascade}). The wall and fix hold across three off-the-shelf memory systems, a
frontier memory-\emph{writer}, real dialogue (MultiWOZ, \citealp{budzianowski2018multiwoz}), and
$\tau$-bench \citep{yao2024taubench}, where they replicate scale-invariantly from $3$B to the frontier
under the benchmark's own database-state scoring (\S\ref{sec:generality},~\ref{sec:realmem}). On
$\tau$-bench's mandatory tool interface, whether a lossy memory becomes a harmful \emph{action} is a
model-by-interface property: the frontier escalates where open models commit the wrong action, and
removing the safe-exit affordance makes all but the strongest commit it too --- a prescription to keep an
explicit escalate/abstain affordance in a tool schema that must fail safe.
\end{itemize}

\section{Related Work}
\label{sec:related}

\paragraph{Memory, retrieval, and context compression.}
Assistant and agent memory systems compress history into carried-forward notes or retrievable chunks
\citep{packer2024memgpt, park2023generative, zhong2024memorybank, lewis2020retrieval}, and a parallel line
compresses the prompt or context itself \citep{jiang2023llmlingua, mu2023gist, chevalier2023autocompressor};
long contexts are in any case attended unevenly, with material in the middle used least
\citep{liu2024lost}. These systems compress toward the salient conclusion. Our result is a caution for
exactly that default: a summary that keeps the conclusion and sheds the source preserves an error and
destroys the means to repair it.

\paragraph{Hallucination and abstention.}
The disposition at the center of our failure, emitting a confident value rather than declining, is the
subject of a large hallucination literature \citep{ji2023hallucination, huang2023hallucination}, whose
remedy is calibrated abstention: a model that knows what it does not know
\citep{kadavath2022know, yin2023know} should decline once its support is gone. We show that whether the
support survives is set upstream, by the compression policy. The same model abstains or confabulates
depending only on what its memory kept, so abstention here is not a fixed trait to be calibrated but an
outcome a memory policy can create or destroy.

\paragraph{Self-correction and feedback.}
A line of work asks models to revise their own outputs
\citep{madaan2023selfrefine, shinn2023reflexion}, and a sober finding is that models often
cannot reliably self-correct reasoning without an external signal
\citep{huang2024selfcorrect}. Our directed correction is precisely such a minimal external
signal, naming the error site but not the fix. We show that even a perfect external signal
is powerless once the recomputable source has left the context: the bottleneck is then
information, not feedback. Nor is our wall a relabeling of anchoring plus the limits of self-correction
\citep{huang2024selfcorrect}. A compression \emph{policy} decides which regime holds at matched budget:
anchoring while the source survives, uncorrectable information loss once it is dropped. And a wrong-valued
memory is \emph{worse} than an empty one, a behavioral asymmetry neither component predicts.

\paragraph{Multi-turn drift.}
That a model can be walked over turns into a state it would not adopt in one shot underlies
multi-turn jailbreaks \citep{russinovich2025crescendo}, and the tendency to follow a confident
interlocutor rather than the evidence is documented as sycophancy \citep{sharma2024sycophancy}. Models
also degrade over multi-turn conversation on their own, prematurely committing to an answer and underusing
the middle of the exchange \citep{laban2025lost}. We use a benign, checkable analogue, a planted
arithmetic or logical premise, and study the return trip: whether and when a committed model can be
brought back. Our failure is distinct from both: the model is not
following an agreeable interlocutor but recomputing from what its memory kept, and it fails
when the recomputable source is gone, not when a human pushes.

\paragraph{Knowledge editing.}
Editing a fact in a model's weights is the parametric counterpart to correcting a belief
\citep{meng2022locating, mitchell2022mend, yao2023editing}; ours is the in-context, per-interaction
counterpart, and it isolates a precondition that weight editing sidesteps, namely that the evidence
justifying the new value still be present.

\paragraph{Brittleness hidden from surface inspection.}
A failure can be invisible to inspection of the surface text yet decisive once context is
restored. \citet{choi2026context} show this for \emph{safety}: aligned models clear
content-level guardrails while the very same behavior is unsafe in context, so the danger is
legible to a context-aware evaluator but not to the text alone. We find the same
surface/structure gap for \emph{correctability}: a carried answer can look intact while whether
it can be repaired depends entirely on what structure the compression kept. In both, the
property that matters, safety or correctability, lives in the context or the retained source,
not in the surface that a content check reads.

\section{Brittle Memory and Reclaim Evaluation}
\label{sec:method}

\subsection{Definition and formalization}

\begin{definition}[Reclaim Rate]
Let $M$ be a model and let a task instance have a unique correct answer $y^{*}$. Drift
induces $M$ to commit to a wrong answer $y' \neq y^{*}$. A compression policy $\pi$ maps the
committed interaction to a carried memory $m$ with budget $B(m)$.\footnote{We operationalize
$B$ as a \emph{character} count throughout (the unit the notes are actually truncated to); the
definition is unit-agnostic and ``budget'' elsewhere in the paper means this character budget.}
A directed
correction $\delta$ names the error locus without supplying $y^{*}$. For a memory integrity
$g \in [0,1]$ governing how much of the source survives, the \emph{Reclaim Rate} is
\begin{equation}
\mathrm{RR}(\pi, g) \;=\; \Pr\!\big[\, M(m^{g}_{\pi}, \delta) = y^{*} \,\mid\, \mathrm{drift} \,\big].
\end{equation}
\end{definition}

\noindent We instantiate three policies at matched budget. \lossy{} keeps the salient
conclusion and sheds the source as $g$ falls (the realistic default). \srcfirst{} keeps the
recomputable source and sheds the re-derivable conclusion. \padded{} is the control: it is
\lossy{} padded with neutral filler to \srcfirst{}'s length or beyond, so it carries
\emph{more} text than the fix while retaining no source.

\paragraph{How the notes are built.}
Each note is templated from three recorded fields of the problem: its line-item \emph{source}
(e.g.\ ``7 notebooks at \$4, 9 pens at \$2''), the planted \emph{premise} (``the pens come to
\$27''), and the committed \emph{conclusion} (``the total is \$55''). This makes ``how much
source survives'' concrete and inspectable rather than a free parameter. At $g\ge0.5$ every
policy keeps the source line items, so the source is present under all of them. The wall opens
at $g<0.5$: \lossy{} drops the line items and keeps the premise and the wrong conclusion (at
$g{=}0.1$, only the conclusion), while \srcfirst{} keeps the line items and drops the
conclusion. \padded{} is \lossy{} extended with topic-neutral filler to at least \srcfirst{}'s
character length, equalizing budget. Source-absence is therefore a property of the string we
emit, checkable by testing the note for the line-item tokens, not an assumption: at $g<0.5$ the
\lossy{} note provably contains none of them.

\textbf{Brittle memory} is the regime
\begin{equation}
\begin{aligned}
\mathrm{RR}(\mathsf{lossy}, g) &\to 0 \ \text{as}\ g \to 0, \ \text{while}\\[2pt]
\mathrm{RR}(\mathsf{source\text{-}first}, g) &\ \text{stays high.}
\end{aligned}
\end{equation}
at equal budget $B$. The content-versus-budget control requires
$\mathrm{RR}(\mathsf{lossy\text{-}padded}, g) \approx \mathrm{RR}(\mathsf{lossy}, g)$ with
$B(\mathsf{lossy\text{-}padded}) \ge B(\mathsf{source\text{-}first})$: if the fix were
merely supplying more text, the padded control would reclaim too.

\subsection{The recoverability principle}
\label{sec:principle}

The definition above says \emph{what} brittle memory is; this subsection says \emph{why} it
happens, in a form that predicts the results rather than restating them. The governing idea is
an asymmetry between an answer and the computation that produced it.

\begin{definition}[Re-derivation basis]
A carried memory $m$ is a \emph{re-derivation basis} for a task instance if (i)~the answer
$y^{*}$ is a deterministic function of $m$ the reader can execute, and (ii)~it is
\emph{closed under correction}: applying the corrected operation named by $\delta$ to $m$
yields the corrected answer, still computable from $m$.
\end{definition}

\noindent The source $x$ (the line items) is a re-derivation basis: $y^{*}\!=\!f(x)$ for the
task operation $f$, and for any corrected operation $f_{c}$ the value $f_{c}(x)$ is still
recoverable from $m$. The bare conclusion $y'$ is not: the only function mapping $y'$ to $y^{*}$
is the constant that already presupposes $y^{*}$, which is not one the reader can execute, so $y'$
fails~(i) in the reader-relative sense of the definition; and it fails~(ii) outright, since a
corrected operation cannot be applied to a scalar whose inputs are gone. $y'$ is a \emph{fixed point}
of the correction operator, not a basis for it. (The condition is thus relative to the operations the
reader can actually run, not to the unrestricted existence of a mathematical function.)

\paragraph{The recoverability principle.}
A directed correction acts on the computation, not the answer, so \emph{reclaim succeeds only
to the extent that the carried memory retains a re-derivation basis.} Keeping the answer while
discarding the basis (\lossy) and keeping the basis while discarding the re-derivable answer
(\srcfirst) are opposite choices at equal budget: the first is uncorrectable by construction,
the second correctable. Brittle memory is then a prediction, not an observation.

To state its reach, let $\pi(m)\in[0,1]$ be the fraction of the re-derivation basis surviving
compression and $\alpha(m)$ its \emph{accessibility} (whether the reader can locate the
surviving basis in $m$). Reclaim needs the basis present \emph{and} findable, and a reader of
capability $\kappa$ to exploit it: heuristically
$\mathrm{RR}\approx\pi(m)\cdot\alpha(m)\cdot s(\pi,\kappa)$, where $s$, the ability to re-derive from
the surviving fragment, \emph{saturates} as
$\pi\!\to\!1$ (a complete basis is trivial to re-derive from, so weak and strong readers both
succeed) and is strongly $\kappa$-dependent at intermediate $\pi$ (reconstructing the answer
from a fragment takes capability). This is a qualitative organizing heuristic, not
a fitted model: we use it to predict the \emph{shape} of every result, not the value of any
cell.

Three regimes follow. At $\pi{=}0$ (\lossy{} at the wall) reclaim collapses to a
\emph{capability-invariant} floor: no basis, nothing capability can exploit. At $\pi{=}1$
(\srcfirst{} with the source intact) reclaim is high and tracks a capability ceiling. In
between ($0<\pi<1$, a partial basis) reclaim is intermediate and maximally
capability-sensitive. The non-obvious consequence, which we name here and confirm below, is
that the benefit of a stronger model is \emph{not monotone} in how much of the basis survives.

\paragraph{Capability inverted-U.}
The gain from capability is near zero when the basis is absent ($\pi{=}0$, nothing to
exploit) or complete ($\pi{=}1$, trivial to exploit), and greatest when the basis is
\emph{partial}: a better model helps most exactly where the memory kept \emph{some} of the
recomputation path. We report this signature in two independent settings: the paired severity sweep
and the deployed memory systems (\S\ref{sec:generality}; Figure~\ref{fig:invertedu}).

Two axes therefore run through the results: \emph{reclaimability} (the recoverability
principle, governed by $\pi$ and $\alpha$, which sets \rr{}) and \emph{safety under failure}
(whether $m$ carries a false anchor $y'$, which sets the emit-versus-abstain response of
\S\ref{sec:results}). The principle governs the first and is deliberately silent on the
second; keeping them apart is what makes each falsifiable.

\subsection{The reclaim protocol}

Each task instance has a known answer, so success is objective and needs no judge. Exact match
reproduces on whatever models exist later; there is no judge model to deprecate. What persists is the
procedure and the information bound it gates. The per-model cells ride on pinned snapshots and are meant
to be re-measured, not read as fixed constants.

\paragraph{Drift and commitment.}
The opening turn states the problem and injects a wrong intermediate value (``a note says
the pens come to \$27''), inducing $M$ to commit to a wrong answer. We deepen the
commitment over up to eight neutral follow-up turns that re-use the wrong figure without
re-deriving the corrupted component, checkpointing the interaction at commitment depths
$\{1,2,4,8\}$.

\paragraph{Correction, two forms.}
At a checkpoint we deliver either a \emph{generic} correction (``something above is wrong,
recheck'') or a \emph{directed} one that names the error locus in the trace's own terms
without giving the answer (``the pens subtotal is wrong, recheck that''). A success is the
model recomputing, not copying. The directed correction is the strongest realistic external signal
short of supplying the answer. We therefore read it as an \emph{upper bound} on locus-naming correction
quality: if even a locus-naming correction cannot reclaim once the source is gone, a vaguer real-world one
cannot either. The idealization is conservative for the wall. (Handing over the value itself is a stronger signal still; that case is
the correction taxonomy of Table~\ref{tab:corrtax}, where \lossy{} fails even then.)

\paragraph{Cross-session compression.}
To study genuine information loss, the corrected query is issued in a fresh session whose
only inheritance is a single carried memory of the first, written under one of the three
policies at integrity $g \in \{1.0, 0.6, 0.3, 0.1\}$. This separates two regimes that look
identical from outside: \emph{anchoring} (the information is present, the model is
entrenched) and \emph{information loss} (the source is gone).

\section{Experimental Setup}
\label{sec:setup}

\paragraph{Tasks.}
We use two families of thirty-two problems each. The first is multi-step arithmetic with a
clean pre-tax total: the conclusion is a deterministic function of a small, fully
recomputable source (the line items). The second is non-arithmetic constraint logic
(role, seating, ordering, and color puzzles) with a single-token answer: the conclusion is
a logical, not numeric, deduction over a clue set. Single-token and numeric answers keep
scoring objective. The headline cross-session cells use all thirty-two problems per family
($n{=}96$ at three seeds); a fixed canonical eight-problem subset ($n{=}24$) is used for the cheaper
runs, the \texttt{grok-4.3} board, the single-conversation window (Table~\ref{tab:single}), and the
size, noise, and completeness sweeps.

\paragraph{Models.}
We evaluate \texttt{llama-3.1-8b-instruct} \citep{metaai2024llama31} and \texttt{grok-4.3}
\citep{xai2026grok} (pinned snapshot), an $8$B open model and a frontier system, both accessed
through OpenRouter's OpenAI-compatible endpoint; the Claude frontier models used in the
answering-model replay (\texttt{claude-sonnet-4-6}, \texttt{claude-opus-4-8}) are accessed through
the Anthropic API. Every condition is run over three seeds at temperature $0.7$ (the Claude models
in the replay run at their default, as Opus does not accept the parameter). The cross-session sweep
on the frontier model is $1{,}224$ calls per task. The disposition breadth sweep
(App.~\ref{app:disposition}) holds the memory fixed and swaps the answering model across eight systems
from four vendors: \texttt{deepseek-chat} \citep{deepseek2024v3}, \texttt{qwen-2.5-7b}
\citep{qwen2024qwen25}, and \texttt{gpt-4o-mini} \citep{openai2024gpt4omini} via OpenRouter,
\texttt{gpt-5.4} \citep{openai2026gpt54} via the OpenAI API, and \texttt{grok-4.3} via the official
xAI API, with the Claude models via the Anthropic API; the reasoning and frontier models (\texttt{gpt-5.4},
\texttt{grok-4.3}, and Claude) run deterministically.

\paragraph{Validators.}
Three checks, run for free against a deterministic fake, can each fail: that the planted
premise actually drifts the model, that the window favors the directed arm, and, as the
central anti-rig check, that when the source's line-item tokens are absent from the carried
note, reclaim fails for \emph{both} arms. The fake reclaims only when those tokens are present
in its context, so a passing run cannot be faked by a model that merely pattern-matches the
correction. All three pass ($3/3$). Because absence is read off the constructed note (a token
test) rather than assumed, the wall rows are a clean $0.00$ by measurement, not stipulation.

\section{Results}
\label{sec:results}

\subsection{The wall, and why it is the setup}

\paragraph{Within one conversation, the window is anchoring, not forgetting.}
Table~\ref{tab:single} reports reclaim on the small model when the full trace remains in
context. A generic correction has a real window that shuts as the model entrenches
(\rr{} $0.42 \to 0.04$ over eight turns), while a directed correction holds far longer
($0.79 \to 0.50$). Counter to a forgetting account, pushing the error \emph{back} behind
unrelated filler \emph{lifts} the generic correction ($0.17 \to 0.50$): the information
never leaves context, so distance cannot starve reclaim, it only loosens the anchor. A
directed correction beats the generic one at every depth and distance. One intact
conversation therefore has no wall, only an anchor that a directed signal overcomes.

\begin{table}[tbp]
\centering\small
\caption{\textbf{The single-conversation window is anchoring} (\texttt{llama-3.1-8b},
eight problems $\times$ three seeds). Reclaim Rate versus commitment depth and versus
unrelated filler distance at fixed deep commitment. Distance lifts generic reclaim, the
opposite of a forgetting wall.}
\label{tab:single}
\begin{tabular}{ccc@{\hskip 1.4em}ccc}
\toprule
depth & gen & dir & dist. & gen & dir\\
\midrule
1 & 0.42 & 0.79 & 0  & 0.17 & 0.62\\
2 & 0.21 & 0.79 & 4  & 0.46 & 0.71\\
4 & 0.58 & 0.71 & 8  & 0.46 & 0.83\\
8 & 0.04 & 0.50 & 16 & 0.50 & 0.79\\
\bottomrule
\end{tabular}
\end{table}

\paragraph{Across sessions, the window becomes a wall.}
When only a lossy memory crosses the session boundary, reclaim holds while the memory keeps
the source and collapses once it is compressed past it (Table~\ref{tab:wall},
\lossy{} columns). Past the threshold the carried note contains a wrong answer and nothing
to check it against, and even a directed correction dies: there is no error site left to
point at. The directed advantage that dominated the single conversation vanishes here, and
its disappearance is itself a diagnostic for the regime change from anchoring to
information loss.

\paragraph{The wall sits in the same place regardless of capability.}
The failure is identical across the two base models we run end to end (\texttt{llama-3.1-8b} and
\texttt{grok-4.3}; Table~\ref{tab:wall}) and across the answering-model replay that swaps the reader
over a fixed memory (\S\ref{sec:generality}, Table~\ref{tab:frontier}). On arithmetic the frontier
model walls at a perfect $0.00$ exactly where the small model does: strictly better wherever
information survives, exactly as helpless where the source was dropped. What the model does once the
source is gone is the finding, measured next.

\subsection{The behavioral finding: a wrong-valued memory is worse than an empty one}

\paragraph{The wall is silent, not safe.}
The deployment-relevant content of the wall is not its height but what the model does \emph{instead}
of recovering (Table~\ref{tab:failmode}), and whether it declines is model-specific. When a model does
answer, it hands back a wrong value rather than recomputing, and that value is overwhelmingly the
inherited attractor. Grok, when it commits, emits the exact attractor and abstains otherwise; the $8$B
model mostly abstains and splits its few emissions between the inherited value and a fresh wrong one.\footnote{These emit-versus-abstain fractions are
answering-temperature-sensitive, and the base models run at $0.7$ while Opus is deterministic, so the
cross-model \emph{spread} is behavior under each model's own decoding, not a pure capability
ordering. A matched greedy-decoding control (the temperature-$0.7$ open models rerun at temperature $0$)
confirms the \emph{disposition} is not a decoding artifact: the strongly-emitting models stay high
(\texttt{deepseek} $0.93{\to}0.84$, \texttt{qwen} $0.99{\to}0.94$), while \texttt{gpt-4o-mini}'s smaller
free-text emission ($0.29$) resolves to a clean abstention ($0.00$ at temperature $0$), consistent with
its abstaining placement. The within-memory source-kept/source-dropped contrast is unaffected regardless:
it holds the carried string fixed and only swaps what the memory kept.} We score the value the model commits on its \texttt{ANSWER} line, not whether the surrounding prose
hedges (\emph{strict scoring}, used for every emission count in this paper, Tables~\ref{tab:blank},
\ref{tab:failmode},~\ref{tab:attractor},~\ref{tab:disposition}): a source-less model emits a wrong
answer rather than flagging that it cannot verify, a failure a downstream system never sees. Strict
scoring is a deliberate, parser-facing choice. Where a model \emph{does} hedge in prose while
committing the wrong value on its answer line, that split is measured directly on MultiWOZ: Opus
caveats ``unverified'' yet still emits the drifted time on half its answer lines
(\S\ref{sec:realmem}), which is precisely the quantity a hedge-crediting scorer would recover. This
behavior, not the information loss, is the result.

The reader's instinct is that any memory beats none. It is exactly backwards, and a matched test
isolates it. At the wall we swap the lossy note for a \emph{blank} one that kept neither source nor
conclusion. The session-2 prompt is otherwise identical: it names the quantity asked and the error
locus but carries no line items and no prior value. Under a blank note the model has nothing to compute
from, so its abstention is close to structural rather than a free behavioral choice; the informative
comparison is what the \emph{identical} prompt does when a stale value is instead present. With nothing
to inherit, both base models abstain on every problem ($0.00$ wrong emission, $n{=}96$). The same models
under lossy compression instead emit a confident wrong value ($0.17$ on llama, $0.57$ on grok, of which
$0.10$ and $0.57$ is the exact inherited attractor), and neither recovers the truth (no source
survives). The gap is purely behavioral: keeping the stale conclusion does nothing for correctability
and turns a safe abstention into a confident error (Table~\ref{tab:blank}). The same holds on the
model's \emph{own} self-generated errors, only milder (the $8$B model mostly abstains), so the planted
note is the more adversarial case, not a special one (\S\ref{sec:limitations}). It holds on real
dialogue too (Table~\ref{tab:attractor}): for the models that answer, a memory that kept the wrong
value is again worse than one that kept nothing.

Across the eight models we test (App.~\ref{app:disposition}) the direction never reverses. No model
emits a confident wrong value \emph{less} often under a lossy memory than an empty one, so a lossy
memory is never better than an empty one \emph{for recovering the drifted answer}. The claim is scoped to
that correction utility: a lossy memory can serve other ends, an approximate estimate or a correlated
value, where exact re-derivation is not the goal. Within it, a lossy memory is strictly \emph{worse}
only where the model is disposed to answer. The gap is large on the models that emit rather than abstain (deepseek $+0.83$,
grok $+0.57$, qwen $+0.39$, llama $+0.17$) and vanishes on the four OpenAI and Anthropic models, which
abstain under both memories (\texttt{gpt-4o-mini}, \texttt{gpt-5.4}, Sonnet, and Opus, all at $0.00$).
That even the small, cost-tier \texttt{gpt-4o-mini} abstains alongside the larger ones shows the split
does not track model scale; across the snapshots we evaluate it aligns with vendor and training family
rather than capability. Abstention is the
escape, and source-first compression removes the need for it by restoring what the model needs to
recompute. That escape depends on the interface offering a decline: through a structured-output field
with no abstain option, as many required tool arguments are, the frontier models that abstain here
instead commit the inherited wrong value. A schema that admits a null or ``insufficient'' value keeps the
decline available, so the hazard is a property of no-abstain field design, not of structured output as
such (\S\ref{sec:interface}).

\begin{table}[tbp]
\centering\footnotesize
\setlength{\tabcolsep}{5pt}
\caption{\textbf{Lossy memory is worse than empty memory on arithmetic} (base models, wall $g{=}0.1$,
directed, $n{=}96$/cell). An \emph{empty} memory makes both models abstain; a \emph{lossy} one that kept
the wrong conclusion makes them emit a confident wrong value, much of it the inherited attractor. Neither
recovers the truth (no source), so the difference is purely behavioral: the wrong-valued memory converts
an abstention into a confident error. Arithmetic counterpart of the MultiWOZ result
(Table~\ref{tab:attractor}).}
\label{tab:blank}
\begin{tabular}{@{}llccc@{}}
\toprule
model & memory & emit wrong & abstain & attractor\\
\midrule
\multirow{2}{*}{\texttt{llama-3.1-8b}} & blank & $0.00$ & $\mathbf{1.00}$ & $0.00$\\
                                       & lossy & $0.17$ & $0.83$ & $0.10$\\
\addlinespace[2pt]
\multirow{2}{*}{\texttt{grok-4.3}}     & blank & $0.00$ & $\mathbf{1.00}$ & $0.00$\\
                                       & lossy & $0.57$ & $0.43$ & $0.57$\\
\bottomrule
\end{tabular}
\end{table}

\subsection{The frontier escape is an interface affordance}
\label{sec:interface}

The escape just established, the frontier models abstain rather than emit a wrong value, 
requires the interface to \emph{offer} a decline. Deployed agents rarely call a model through free text; they use
structured outputs and tool calls whose answer field is frequently mandatory. We rerun the wall
($g{=}0.1$, directed, $n{=}96$/cell) through two structured channels, holding the drifted memory, the
directed correction, and the strict scoring fixed, changing only the answer channel. In the \emph{soft}
channel the model returns a JSON object with an \texttt{answer} field it may set to
\texttt{"INSUFFICIENT"}; in the \emph{hard} channel that field is a required number with no abstain token
(a mandatory tool argument).

The soft channel leaves the escape intact but shows the emission rate is itself an interface artifact,
and a \emph{non-monotone} one. An explicit abstain field \emph{raises} \texttt{llama}'s emission
($0.17\!\to\!0.48$: the prose channel had hidden commitments behind hedging) and \emph{lowers}
\texttt{grok}'s ($0.57\!\to\!0.00$: the field is an easier decline than prose), while the frontier
models stay at $0.00$. Whether prose or a field better elicits a hedge is model-specific, so the exact
emission fraction is not a stable model property.

The hard channel removes the escape entirely (Table~\ref{tab:interface}). Forced to fill the field, every
model commits a value on every lossy problem, and that value is overwhelmingly the inherited attractor:
$0.49$ on \texttt{llama}, $0.72$ on \texttt{gpt-5.4}, $0.81$ on \texttt{gpt-4o-mini}, $0.85$ on Opus,
and $1.00$ on \texttt{grok} and Sonnet. The frontier models that walled at a perfect $0.00$ in free text now emit the
stale wrong value $0.72$--$1.00$ of the time. The matched \emph{blank} memory is the control that makes
this specific: with nothing to inherit, a forced guess equals the exact attractor $0.00$ of the time on
every model. The lossy commitment is the poison, not the forcing. Sonnet draws the line sharply: it
still abstains on half the \emph{blank}-hard problems, declining to invent a value from nothing even
under a mandatory field, yet commits the inherited value on \emph{all} of the lossy ones. The
disposition to abstain that escaped brittle memory above is thus an affordance of the free-text
interface. Under the mandatory-field interfaces production agents actually use, a lossy memory poisons a
frontier model's output most or all of the time. A blank memory forced through the same field is not
``clean'' either, it too emits a wrong number, but an \emph{arbitrary} one that matches the
planted attractor $0.00$ of the time. What lossy adds is a \emph{systematic, inherited} error rather
than a random one: the kind that propagates coherently through a memory loop where a random guess does
not.

\begin{table}[tbp]
\centering\footnotesize
\setlength{\tabcolsep}{5pt}
\caption{\textbf{The frontier escape is an interface affordance} (wall $g{=}0.1$, directed,
$n{=}96$/cell). \emph{Prose} and \emph{soft} give confident-wrong emission under lossy memory: the frontier
models escape at $0.00$. \emph{Hard} makes the answer field a mandatory number with no abstain token (as a
required tool argument), so every model emits; we then report the \emph{inherited-attractor} rate, whether
the forced value equals the stale wrong value, under lossy versus the matched blank memory. The escape
vanishes: every model commits the inherited value under lossy ($0.49$--$1.00$) and none under blank
($0.00$), so the abstention was an interface affordance, not a disposition.}
\label{tab:interface}
\begin{tabular}{@{}lcccc@{}}
\toprule
 & \multicolumn{2}{c}{lossy emit wrong} & \multicolumn{2}{c}{hard: attractor} \\
\cmidrule(lr){2-3}\cmidrule(lr){4-5}
model & prose & soft & lossy & blank \\
\midrule
\texttt{llama-3.1-8b} & $0.17$ & $0.48$ & $0.49$ & $\mathbf{0.00}$ \\
\texttt{grok-4.3}     & $0.57$ & $0.00$ & $1.00$ & $\mathbf{0.00}$ \\
\texttt{gpt-4o-mini}  & $0.00$ & $0.00$ & $0.81$ & $\mathbf{0.00}$ \\
\texttt{gpt-5.4}      & $0.00$ & $0.00$ & $0.72$ & $\mathbf{0.00}$ \\
Sonnet                & $0.00$ & $0.00$ & $1.00$ & $\mathbf{0.00}$ \\
Opus                  & $0.00$ & $0.00$ & $0.85$ & $\mathbf{0.00}$ \\
\bottomrule
\end{tabular}
\end{table}

\subsection{The source-first remedy}

\paragraph{The lever is content, not budget.}
The \padded{} control isolates content from length: it carries more text than \srcfirst{} and still
walls identically to plain \lossy{} ($0.00$ on arithmetic; Table~\ref{tab:wall}). At equal or greater
budget, the policy that keeps the source reclaims and the policy that keeps the conclusion fails, so
the lever is what the memory keeps, not how much.

\paragraph{The wall is a choice, removed by source-first in the regime that matters.}
\srcfirst{}'s advantage is a low-integrity effect, and a decisive one. Where lossy compression has
shed the source, \srcfirst{} reclaims ($0.99$--$1.00$ on arithmetic) while both lossy variants wall
($0.00$), with non-overlapping $95\%$ intervals on both models (Tables~\ref{tab:wall},~\ref{tab:logic}).
At high integrity, where lossy still carries the source, the policies are comparable and sometimes
reverse: \padded{} exceeds \srcfirst{} at $g{=}1.0$ on llama arithmetic ($0.85$ vs $0.61$) and at
$g{=}0.6$ on grok logic. That reversal is likely a positioning effect rather than text aiding recovery,
but it is a real cost of dropping the conclusion when the source would have survived anyway, which is
exactly why we scope the prescription to budget pressure and recommend keeping both where the budget
fits (below). So we claim no all-regime dominance, and we do not rest the content-not-text claim on
these cells; the matched control at the wall, where the padding carries \emph{no} source, carries it.
The low-integrity result is a rule, not a coincidence, because a conclusion is a deterministic function
of its source: a kept source regenerates the conclusion, a kept conclusion never regenerates the source.
Under budget pressure, keeping the source is what stays correctable.

\paragraph{Dropping the conclusion is not only for budget.}
Source-first drops the conclusion to make room for the source, but the conclusion is also a
liability. A note that keeps both at the wall reclaims $0.95_{[.90,.99]}$ on the $8$B reader, below
source-first's $1.00$: the stale conclusion re-attracts a weak reader even with the source beside it.
On the frontier readers the cost vanishes ($1.00$ on Sonnet and Opus). So dropping the conclusion is
safe under budget pressure: it removes a weak-reader trap and costs nothing on a strong reader. When
the budget fits both, keeping the conclusion is also fine on a strong one, and lets it cross-check by
recomputing. Either way, the source is the load-bearing part.

This also explains \srcfirst{}'s apparent non-monotonicity in Table~\ref{tab:wall} (llama arithmetic
$0.61$ at $g{=}1.0$ rising to $0.99$ at $g{=}0.1$). At high integrity ($g{\ge}0.5$) the \srcfirst{}
note by construction carries the planted premise \emph{and} the stale conclusion alongside the source
(\S\ref{sec:method}), so a weak reader re-anchors on the poison it also kept. Dropping both below $0.5$
is what leaves \srcfirst{} cleanest exactly at the wall, where correctability matters. More of the
interaction preserved is not more correctable when the extra is the liability, which is the point.

\subsection{Robustness to corrections}
\label{sec:robustness}

A deployed correction is rarely a clean locus, so the fix must survive weaker and adversarial ones. It
does, because the work is done by the restored source. Under a \emph{vague} ``something is off'' nudge,
\srcfirst{} reclaims essentially as well as under a directed locus ($\approx1.00$ on arithmetic,
$0.76$--$0.82$ on logic) while \lossy{} stays at its floor: specificity governs in-context anchoring but
is irrelevant once the lever is the restored source. It is equally robust to corrections that
\emph{mislead}. A \emph{false locus} naming a correct component as the error leaves \srcfirst{} at the
true value ($1.00$ on both base models, $n{=}24$ each). A \emph{confident wrong value} asserted as
established fact, the sycophancy case \citep{sharma2024sycophancy}, is rejected on every trial on
both frontier readers; \lossy{} there abstains rather than recompute, adopting the asserted value on
only $0.09$ of Sonnet trials and $0.03$ of Opus. The adversarial members are capability-gated. Against a
\emph{sustained} four-turn push and an injected \emph{fabricated source}, frontier readers resist
($0.90$--$1.00$) while the $8$B reader caves ($0.00$--$0.27$), the same capability ladder that
governs reading a fuzzy source. The full battery, with the false-locus and
injection mechanics and Tables~\ref{tab:generic} and~\ref{tab:adversarial}, is in
App.~\ref{app:robustness}.

Two endpoints complete the taxonomy (Table~\ref{tab:corrtax}). A content-free ``are you sure?'' makes
\srcfirst{} recompute, and makes \lossy{} only re-assert the wrong value or abstain. Even the
\emph{correct value}, handed over as an instruction, fixes \lossy{} on the smaller models but only
partly on Opus, which commits the corrected value on $0.22$ of trials and abstains on the rest, never
re-asserting the stale one. The frontier reader thus declines rather than defends: stripped of the
source it withholds an answer instead of either recomputing or confidently re-emitting the stale value.
So \lossy{} is correctable by none of these short of the answer, and only partly even then, while
\srcfirst{} is correctable across the spectrum.

\begin{table}[tbp]
\centering\small
\caption{\textbf{Two correction types complete the taxonomy} (arith wall $g{=}0.1$, fraction returning the
\emph{true} value; llama/Sonnet $n{=}96$, Opus $n{=}32$). A content-free ``are you sure?'' makes
\srcfirst{} recompute and only makes \lossy{} re-assert the wrong value. Even an \emph{explicitly supplied
correct value} fixes \lossy{} on the smaller models but only partly on Opus, which commits the corrected
value on $0.22$ of trials and abstains on the rest, never re-asserting the stale one.}
\label{tab:corrtax}
\begin{tabular}{@{}llccc@{}}
\toprule
the user says & policy & llama & Sonnet & Opus\\
\midrule
``are you sure?''   & \srcfirst{} & $1.00$ & $1.00$ & $1.00$\\
                    & \lossy{}    & $0.00$ & $0.00$ & $0.00$\\
\midrule
the correct value   & \srcfirst{} & $1.00$ & $1.00$ & $1.00$\\
                    & \lossy{}    & $1.00$ & $1.00$ & $\mathbf{0.22}$\\
\bottomrule
\end{tabular}
\end{table}

\subsection{The wall's severity, hard and soft}
\label{sec:severity}

\paragraph{The failure's severity is conditional, and the same on both models.}
The wall is a clean $0.00$ on arithmetic, where lossy compression discards the actual
numbers and nothing can be reconstructed, and \emph{soft} on logic
(Table~\ref{tab:logic}), where the lossy note retains a corrupted relational clue in a small
constraint space. Because the logic answer is one of a few tokens, we measure the free-guess floor directly: a blank
note giving only the candidate set (no clue, no conclusion) reclaims at $0.04$--$0.17$
(App.~\ref{app:chance}), below the $\approx0.30$ uniform rate, the models abstain or anchor rather
than guess freely. Against that measured floor the soft wall separates by capability. On the frontier
model the surviving clue lifts directed reclaim to $0.42$--$0.50$, far above its $0.12$ guess floor.
That is genuine re-derivation, and it is the reason the floor is not circular: were reclaim merely
scoring its own setup, logic would zero out like arithmetic. On the small model reclaim reaches only
$0.16$ (and $0.03$ recovery in the failure-mode count), at or below its own $0.17$ floor, scarcely
distinguishable from guessing. The soft wall is therefore real
re-derivation for a capable reader and chance for a weak one. \srcfirst{} keeps all of the
source either way.

\begin{figure}[t]
\centering
\includegraphics[width=0.60\textwidth]{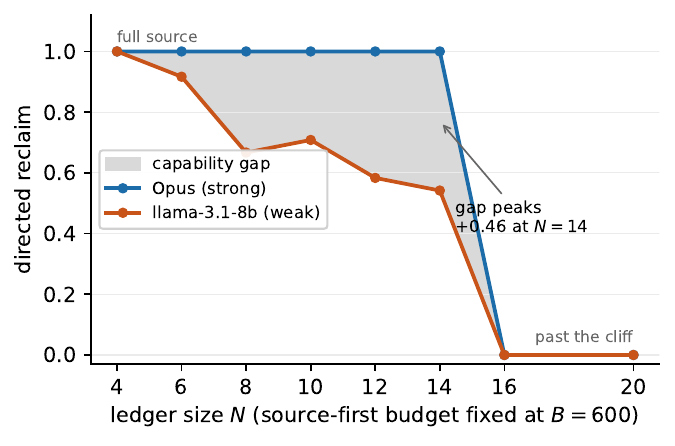}
\caption{\textbf{The capability inverted-U, swept within arithmetic.} Directed reclaim of a strong
(Opus) and a weak (\texttt{llama-3.1-8b}) reader under \srcfirst{}, against ledger size $N$ at a fixed
carried-memory budget ($B{=}600$). Growing $N$ at fixed budget drives the note from full source
($k{=}N$ items kept, left) through partial ($k<N$) to past the cliff (right), a graded partial-source
regime on a single arithmetic task. The capability gap (shaded) traces the inverted-U: near zero at
full source (both reclaim), peaking at $+0.46$ where the source is partial (the strong reader
re-derives from the fragment, the weak one cannot), and near zero past the cliff (both wall, nothing to
exploit). This is a within-task measured curve, not a categorical ordering, and it recovers the
three-regime prediction of \S\ref{sec:principle} with the intermediate peak the principle requires. The
same non-monotonicity holds independently across the deployed memory systems (\S\ref{sec:generality}).}
\label{fig:invertedu}
\end{figure}

\paragraph{The capability gap traces the inverted-U of \S\ref{sec:principle}.}
We measure this within arithmetic as a swept curve (Figure~\ref{fig:invertedu}): holding the
\srcfirst{} budget fixed and growing the ledger degrades the source from full through partial to past
the cliff, and the Opus$-$llama gap rises from $0.00$ at full source to $+0.46$ where the source is
partial, then falls back to $0.00$ once nothing remains. The same non-monotone shape appears
categorically across the three regimes: arithmetic \lossy{} ($\pi{=}0$, both wall, gap $0.00$),
\srcfirst{} ($\pi{=}1$, both reclaim, gap $0.01$), and the soft logic wall ($0<\pi<1$, a capable reader
re-derives from the surviving fragment while a weak one cannot, gap $\approx0.3$; Table~\ref{tab:logic}). The gap is largest where the memory kept \emph{some} of the
recomputation path, as \S\ref{sec:principle} predicts, and it recurs across the deployed systems
(\S\ref{sec:generality}).

\begin{table}[tbp]
\centering\footnotesize
\setlength{\tabcolsep}{3.5pt}
\caption{\textbf{Arithmetic: the wall (\lossy), the length control (\padded), and the fix (\srcfirst), two
models.} Directed \rr{} ($95\%$ CI) vs.\ memory integrity $g$ (llama $n{=}96$; grok $n{=}24$;
temperature $0.7$). At low integrity (the wall) \lossy{} and \padded{} sit at $0.00$ while \srcfirst{} is
$0.99$--$1.00$, non-overlapping. At high integrity \lossy{} still keeps the source and the policies are
comparable; the fix is a low-integrity effect, the regime budget pressure produces.}
\label{tab:wall}
\begin{tabular}{lcccccc}
\toprule
 & \multicolumn{3}{c}{\texttt{llama-3.1-8b}} & \multicolumn{3}{c}{\texttt{grok-4.3}}\\
\cmidrule(lr){2-4}\cmidrule(lr){5-7}
$g$ & \lossy & \padded & \srcfirst & \lossy & \padded & \srcfirst\\
\midrule
1.0 & $0.61_{[.52,.71]}$ & $0.85_{[.78,.92]}$ & $0.61_{[.52,.71]}$ & $1.00_{[1,1]}$ & $1.00_{[1,1]}$ & $1.00_{[1,1]}$\\
0.6 & $0.82_{[.74,.90]}$ & $0.85_{[.78,.92]}$ & $0.70_{[.60,.78]}$ & $1.00_{[1,1]}$ & $1.00_{[1,1]}$ & $1.00_{[1,1]}$\\
0.3 & $0.01_{[0,.03]}$ & $0.00_{[0,0]}$ & $\mathbf{0.99_{[.97,1]}}$ & $0.00_{[0,0]}$ & $0.00_{[0,0]}$ & $\mathbf{1.00_{[1,1]}}$\\
0.1 & $0.00_{[0,0]}$ & $0.00_{[0,0]}$ & $\mathbf{0.99_{[.97,1]}}$ & $0.00_{[0,0]}$ & $0.00_{[0,0]}$ & $\mathbf{1.00_{[1,1]}}$\\
\bottomrule
\end{tabular}
\end{table}

\begin{table}[tbp]
\centering\footnotesize
\setlength{\tabcolsep}{3.5pt}
\caption{\textbf{Constraint logic: the same three policies, two models.} Directed \rr{} ($95\%$ CI) vs.\
memory integrity $g$ (llama $n{=}96$; grok $n{=}24$). At low integrity \srcfirst{} beats both lossy
variants; the wall is \emph{soft} rather than a clean zero, because the lossy logic note leaves a
reconstructable clue. At high integrity the policies trade places, so as on arithmetic the fix is a
low-integrity effect.}
\label{tab:logic}
\begin{tabular}{lcccccc}
\toprule
 & \multicolumn{3}{c}{\texttt{llama-3.1-8b}} & \multicolumn{3}{c}{\texttt{grok-4.3}}\\
\cmidrule(lr){2-4}\cmidrule(lr){5-7}
$g$ & \lossy & \padded & \srcfirst & \lossy & \padded & \srcfirst\\
\midrule
1.0 & $0.52_{[.43,.61]}$ & $0.36_{[.27,.46]}$ & $0.46_{[.36,.56]}$ & $0.71_{[.50,.88]}$ & $0.96_{[.88,1]}$ & $0.75_{[.58,.92]}$\\
0.6 & $0.40_{[.30,.50]}$ & $0.38_{[.28,.48]}$ & $0.50_{[.40,.59]}$ & $0.92_{[.79,1]}$ & $1.00_{[1,1]}$ & $0.83_{[.67,.96]}$\\
0.3 & $0.16_{[.08,.23]}$ & $0.18_{[.10,.26]}$ & $\mathbf{0.76_{[.68,.84]}}$ & $0.42_{[.21,.62]}$ & $0.38_{[.21,.58]}$ & $\mathbf{0.92_{[.79,1]}}$\\
0.1 & $0.05_{[.01,.10]}$ & $0.09_{[.04,.16]}$ & $\mathbf{0.79_{[.71,.86]}}$ & $0.50_{[.29,.71]}$ & $0.50_{[.29,.71]}$ & $\mathbf{0.96_{[.88,1]}}$\\
\bottomrule
\end{tabular}
\end{table}

\begin{table}[tbp]
\centering\small
\caption{\textbf{What the model does when reclaim fails}, at the wall (\lossy{}, $g{=}0.1$, directed;
$n{=}96$/row except grok logic, $n{=}32$). \emph{recov.}\ correct despite the lossy memory; \emph{inherit}\
repeats the compressed wrong value; \emph{novel}\ a different wrong value; \emph{abst.}\ declines.
Arithmetic recovers nothing; the model either inherits the stale value or abstains (grok inherits more
often, the $8$B model mostly abstains). Logic recovers $3$--$53\%$ from a surviving clue, so its wall is
soft on the frontier model and near-hard on the $8$B one. The wall rows are the source-absent notes, so
these are failures of \emph{information}, not reasoning (examples, App.~\ref{app:examples}).}
\label{tab:failmode}
\begin{tabular}{llcccc}
\toprule
task & model & recov. & inherit & novel & abst.\\
\midrule
arith & \texttt{grok-4.3} & 0\% & \textbf{57\%} & 0\% & 43\%\\
arith & \texttt{llama-8b} & 0\% & 10\% & 7\% & 83\%\\
logic & \texttt{grok-4.3} & \textbf{53\%} & 9\% & 0\% & 38\%\\
logic & \texttt{llama-8b} & 3\% & 6\% & 1\% & 90\%\\
\bottomrule
\end{tabular}
\end{table}

\section{Generality: Deployed and Frontier Systems}
\label{sec:generality}

\paragraph{Three deployed paradigms, three ways to lose the source.}
The effect is not an artifact of our hand-built notes; it appears in memories people ship,
by several distinct mechanisms. These are general-purpose recall tools, not configured for
correctability; the claim is not that they are poorly built but that their \emph{default}
compression, toward the salient conclusion, sheds the source. We run three off-the-shelf systems
unmodified over the same
session~1 trajectory, each \emph{constructing} its memory with \texttt{llama-3.1-8b}, and
carry that memory into session~2 in the slot our notes occupied: LangChain's
\texttt{ConversationSummaryMemory} \citep{langchain} (a running LLM summary), \texttt{mem0}
\citep{mem0} (LLM fact-extraction
into a retrieved store), and naive vector retrieval over the session's turns.\footnote{FastEmbed
\texttt{bge-small-en-v1.5} \citep{xiao2023cpack} ($384$-dim); one chunk per conversational turn; top $k{=}4$ by
cosine similarity to a fixed correction-shaped query, carried in chronological order. We label
it ``naive'' deliberately: a tuned RAG pipeline could retrieve the source-bearing turns, which
is the point, retrieval helps only when it is aimed at the source, not the conclusion.} (The
writer is the small model; whether a frontier writer would preserve more source is the open
question we flag in Limitations.)
All three wall well below the fix (Table~\ref{tab:frontier}), and each loses the source a
\emph{different} way. The summary \emph{drops} it, compressing toward the conclusion and shedding the
line items. mem0 \emph{buries} it: it keeps the source as extracted facts but bloats each memory with,
on average, $38.1$ numbers absent from session~1 ($100\%$ of its memories carry at least one, against
zero for the summary, naive retrieval, and both hand-built notes; an objective count, no judge). Some
are outright fabrications, such as a made-up pen count introduced to reconcile the planted error; others
are correct re-derivations. Either way, the few line items that actually determine the answer sit in a
mass of generated figures. Naive retrieval \emph{misses} it: keyed on the correction, it surfaces the
conclusion-bearing turns rather than the source-bearing ones, and is the weakest of the three. All cells are $n{=}96$; on the small model, \srcfirst{}'s $95\%$ bootstrap
interval excludes every deployed system on arithmetic.

Aiming the retrieval at the source does not rescue it. A \emph{source-keyed} query over
the same store (asking for the original line items and quantities rather than for the correction)
retrieves different turns but reclaims no better than the conclusion-keyed one
($0.09_{[0,.22]}$ vs.\ $0.06_{[0,.16]}$, overlapping, both far below the distilled
\srcfirst{} note's $0.97_{[.91,1]}$; $n{=}32$, same trajectories). The reason is visible in the retrieved text: in a realistic drift
dialogue the recomputable line items are stated \emph{once} and then buried under turns that
restate and confirm the (wrong) total, so neither query reliably surfaces them. Retrieval direction
is necessary but not sufficient; the source must be salient enough to retrieve, which a distilled
source-first note guarantees and raw retrieval does not. The lever is distillation, not just the
query.

\paragraph{The fix deploys, not just hand-built.}
\srcfirst{}-auto, a one-prompt policy that compresses an arbitrary transcript toward its
recomputable source and away from the conclusion (prompt verbatim in App.~\ref{app:repro}),
beats mem0 and naive retrieval decisively on both tasks ($\ge0.39$ everywhere) and beats the
LangChain summary on arithmetic, where the summary is the closest competitor: the
\srcfirst{}-auto\,$-$\,summary paired difference is $+0.18/+0.32/+0.30$ across Llama/Sonnet/Opus,
each with a $95\%$ bootstrap interval that excludes zero ($[.05,.30]$, $[.22,.43]$, $[.19,.42]$).
On logic the two are \emph{indistinguishable}: the paired difference is within $\pm0.03$ on every
model and every interval straddles zero, because a running summary that happens to retain the ordering
constraints recomputes about as well as an explicit source-first note. Against a good summary, then, the
deployable fix adds essentially nothing on logic. Its value is concentrated on compact numeric sources,
where the summary sheds the line items, and absent where the summary already carries the constraints;
its edge over the fact-extraction and retrieval systems is decisive throughout. The gap to the oracle is
where the idealization shows. The hand-built note perfectly identifies the answer-determining source and
keeps exactly it; \srcfirst{}-auto must \emph{find} that source itself. Its $0.49$--$0.88$, not the
oracle's $1.00$, is the deployable number. On logic this gap is the distiller's, not the reader's:
\srcfirst{}-auto is flat across capability ($0.49/0.53/0.53$ on Llama/Sonnet/Opus) while the hand note
climbs ($0.77\to1.00$), so a stronger reader cannot rescue a logic source the auto-distiller did not
keep. The ``ties the summary on logic'' story is bounded by the distillation prompt, not by the reader.
The axis is source recoverability, not transcript size: a small clean source can sit inside a long
transcript one must compress.

\paragraph{The wall holds on frontier models, and the gap widens.}
We hold each memory fixed and replay only the session-2 answering model across a wide
capability range, an 8B open model up to a frontier reasoner: \texttt{llama-3.1-8b},
\texttt{claude-sonnet-4-6} \citep{anthropic2026sonnet}, and
\texttt{claude-opus-4-8} \citep{anthropic2026opus}, the last a current frontier model widely used for agentic work
(Table~\ref{tab:frontier}). Two facts. Wherever the memory kept the source, capability lifts
reclaim, to a \emph{perfect} $1.00$ on Opus for \srcfirst{}. Wherever the memory dropped it,
every model scores $0.00$, Opus included: lossy compression and naive retrieval are
uncorrectable at any capability we tested. The strongest model therefore has the
\emph{widest} gap between source-kept and source-dropped. The widening is carried by logic,
where source-kept climbs $0.77 \to 1.00$ across the range while source-dropped holds at
$0.00$; on arithmetic source-kept is already at ceiling on the small model ($1.00$), so
there the effect is the floor staying at $0.00$ rather than the ceiling rising. Either way, capability does not move the source-dropped rows, exactly as the hand-built
small-vs-frontier comparison (Tables~\ref{tab:wall},~\ref{tab:logic}) predicts, now on deployed
memories and recognizable models. The capability story thus has two halves. Source-dropped rows holding
at $0.00$ is the near-analytic half: no reader recomputes from an absent string. The substantive half is
source-kept reclaim \emph{climbing} with capability, a stronger reader reads more from a present
source but cannot conjure an absent one. Confabulation is the failure capability helps least with: mem0
stays well below the fix on every model ($0.09/0.12/0.11$ on arithmetic, flat across the capability
range) and gains far less from a stronger reader than the source-bearing systems do, since reading skill
does not, on its own, recover line items buried under the extractor's own figures.

\paragraph{The same inverted-U appears across deployed systems.}
The capability spread (Opus$-$Llama) again peaks at partial retention: near zero where the memory keeps
the source (\srcfirst{}, $0.02$) or drops it entirely (\lossy{}/vector, $\le\!0.08$), and widest for the
running summary that keeps it only \emph{sometimes} ($0.38\!\to\!0.75$, spread $0.37$) and the stochastic
auto-distiller ($0.67\!\to\!0.96$). It is the signature of \S\ref{sec:principle}, now on
independently-built production memories rather than one hand-built sweep.

\begin{table}[tbp]
\centering\footnotesize
\setlength{\tabcolsep}{3.5pt}
\caption{\textbf{The deployed-system wall holds across a wide answering-model range; capability does not
close the source-kept/source-dropped gap.} Directed \rr{} ($95\%$ CI, $n{=}96$/cell) for six session-2
memories held fixed while only the answerer is swapped (llama $\to$ Sonnet $\to$ Opus). Source-kept rows
climb toward $1.00$ on Opus; source-dropped rows (\lossy{}, naive retrieval) stay at $0.00$ on every
model. mem0 stays low throughout, gaining least from capability because the source is buried under
generated figures a stronger reader cannot reliably sort.}
\label{tab:frontier}
\begin{tabular}{l ccc c ccc}
\toprule
 & \multicolumn{3}{c}{arithmetic \rr{} (dir)} & & \multicolumn{3}{c}{logic \rr{} (dir)}\\
\cmidrule(lr){2-4}\cmidrule(lr){6-8}
session-2 memory & Llama & Sonnet & Opus & & Llama & Sonnet & Opus\\
\midrule
\srcfirst{} (hand) & $1.00_{[1,1]}$ & $1.00_{[1,1]}$ & $\mathbf{1.00_{[1,1]}}$ &  & $0.77_{[.69,.85]}$ & $0.90_{[.83,.95]}$ & $\mathbf{1.00_{[1,1]}}$ \\
\srcfirst{}-auto (fix) & $0.53_{[.44,.64]}$ & $0.88_{[.80,.94]}$ & $0.86_{[.79,.93]}$ &  & $0.49_{[.40,.59]}$ & $0.53_{[.43,.64]}$ & $0.53_{[.44,.64]}$ \\
LangChain summary & $0.35_{[.26,.45]}$ & $0.55_{[.45,.66]}$ & $0.56_{[.46,.66]}$ &  & $0.47_{[.38,.57]}$ & $0.53_{[.43,.64]}$ & $0.56_{[.46,.66]}$ \\
\texttt{mem0} & $0.09_{[.04,.16]}$ & $0.12_{[.06,.20]}$ & $0.11_{[.05,.18]}$ &  & $0.10_{[.05,.17]}$ & $0.15_{[.08,.22]}$ & $0.10_{[.05,.17]}$ \\
naive vector retrieval & $0.04_{[.01,.08]}$ & $0.07_{[.02,.12]}$ & $0.07_{[.02,.12]}$ &  & $0.04_{[.01,.08]}$ & $0.03_{[0,.07]}$ & $0.03_{[0,.07]}$ \\
\lossy{} (hand) & $0.00_{[0,0]}$ & $0.00_{[0,0]}$ & $\mathbf{0.00_{[0,0]}}$ &  & $0.05_{[.01,.10]}$ & $0.00_{[0,0]}$ & $\mathbf{0.00_{[0,0]}}$ \\
\bottomrule
\end{tabular}
\end{table}

\paragraph{The deployed wall is the policy, not the writer.}
The deployed memories above are all \emph{written} by \texttt{llama-3.1-8b}, so part of the wall
could be a weak-writer artifact. Re-running memory construction with a frontier writer
(\texttt{claude-sonnet-4-6}), holding the trajectory and the llama answerer fixed, splits the two
LLM-written systems in \emph{opposite} directions: LangChain's summary wall is largely a weak-writer
artifact (a capable writer keeps the line items, lifting directed reclaim $0.38\to0.88$ on
arithmetic), while mem0's is not (a frontier extractor does not rescue it and \emph{confabulates
more}). Writer strength is therefore not a paradigm-independent remedy, and the templated \lossy{}
wall holds throughout, confirming the mechanism is the policy, not the writer (full sub-study,
App.~\ref{app:writer}).

\paragraph{The wall and fix hold end-to-end on a frontier model.}
The frontier results above replay a fixed (llama-written) memory under a stronger reader, isolating
the reading step. We also run the full pipeline frontier-to-frontier: \texttt{claude-sonnet-4-6} both
\emph{writes} the memory and answers, over the $32$ arithmetic problems. The wall holds (templated
\lossy{} $0.00$ directed) and the oracle fix holds (\srcfirst{} $1.00$); a frontier-written LangChain
summary lands at $0.91$. The deployable distiller is the result: \srcfirst{}-auto reaches $1.00$ when
the frontier model writes it, against $0.88$ when llama writes it and the frontier only reads
(Table~\ref{tab:frontier}). The $0.49$--$0.88$ deployable gap to the oracle is therefore largely a
weak-writer artifact, closed by a capable writer, so the shippable number rises with the writer's
capability and not only the reader's.

\paragraph{The wall reproduces as an agentic action.}
The failure is not confined to question answering. In a minimal agentic setting (Battleship), a model
that fires at a grid across turns and loses its own shot record refires an already-fired cell
$0.77$--$0.91$ of the time on the frontier models and never abstains, a silent \emph{action} rather
than a silent answer, and its hit rate collapses ($6.75\!\to\!0.375$ on Opus). Full setup, the
suggestive capability left-edge, and the per-model table are in App.~\ref{app:battleship}.

\paragraph{The wall and fix replicate on a real deployed-agent benchmark, scale-invariantly and judge-free.}
The deployed tests above write their own memories; we close the loop on a benchmark built by others.
$\tau$-bench \citep{yao2024taubench} scores a customer-service agent by comparing the final database state
against an annotated goal state, so correctness is judge-free by construction. We port reclaim onto its
retail domain: for each single-item exchange/modify task the load-bearing source is the requested option
spec (recomputable from the product catalog), and the salient conclusion is the committed exchange. A
writer compresses the case under our four policies; a fresh session receives only that memory plus a
correction and proposes an action, scored by $\tau$-bench's own state hash ($12$ cases $\times$ policies
$\times\,\{$generic, directed$\}\times$ seeds). Table~\ref{tab:taubench} runs this across an eight-model
ladder from a $3$B open model to the current frontier. The wall is present and \emph{scale-invariant}: a
\lossy{} memory recovers the correct action essentially never (pooled \rr{} $0.06$, $\leq 0.14$ at every
scale) and the length-matched \padded{} control tracks it ($0.04$), while \srcfirst{} recovers it (pooled
$0.76$), the recovery rate tracking reader capability exactly as the inverted-U predicts, weak on the left
edge (\texttt{llama-3.2-3b} $0.63$, \texttt{gemini-3.5-flash} $0.26$) and saturating at $1.00$ on the
strong models. The failure is now a wrong \emph{action} on real records, scored without a judge, and the
one-line source-first policy is what removes it. This benchmark isolates the wall and its fix, not the
\emph{worse-than-empty} asymmetry: weaker models commit a wrong action even from a \textsf{blank} memory
(\texttt{llama-3.2-3b} at $0.92$), so \lossy{} is not strictly worse than empty for them, and no model
parrots the stale item id; the emit-versus-abstain result stays an arithmetic and MultiWOZ finding
(\S\ref{sec:results},~\ref{sec:realmem}).

\begin{table}[tbp]
\centering\small
\caption{\textbf{The wall and source-first fix replicate on $\tau$-bench \citep{yao2024taubench},
scale-invariantly and judge-free.} Reclaim Rate (the corrected action reproduces the goal database state,
scored by $\tau$-bench's own state hash) by memory policy, across a $3$B-to-frontier ladder. A \lossy{}
memory recovers the action essentially never at any scale; the length-matched \padded{} control tracks it;
\srcfirst{} recovers it, with the rate tracking reader capability (the inverted-U left edge at $3$B and
Gemini-Flash). Retail domain, single-item exchange/modify cases.}
\label{tab:taubench}
\begin{tabular}{@{}l cccc c@{}}
\toprule
\textbf{Model} & \srcfirst{} & \lossy{} & \padded{} & \textsf{blank} & $n$ \\
\midrule
\texttt{llama-3.2-3b}     & 0.63 & 0.03 & 0.01 & 0.00 & 69 \\
\texttt{llama-3.1-8b}     & 0.96 & 0.14 & 0.14 & 0.22 & 72 \\
\texttt{llama-3.3-70b}    & 1.00 & 0.00 & 0.00 & 0.00 & 72 \\
\texttt{gpt-4o-mini}      & 0.60 & 0.10 & 0.07 & 0.00 & 72 \\
\texttt{gemini-3.5-flash} & 0.26 & 0.11 & 0.04 & 0.00 & 72 \\
\texttt{grok-4.3}         & 1.00 & 0.00 & 0.00 & 0.00 & 72 \\
\texttt{claude-opus-4.8}  & 0.83 & 0.00 & 0.00 & 0.00 & 24 \\
\texttt{gpt-5.4}          & 1.00 & 0.00 & 0.00 & 0.00 & 24 \\
\midrule
\textbf{pooled}           & \textbf{0.76} & \textbf{0.06} & \textbf{0.04} & \textbf{0.03} & 477 \\
\bottomrule
\end{tabular}
\end{table}

\paragraph{At the wall, harm or escalation is a model-by-interface property.}
The run above let the model emit an empty action (abstain); production tool-calling often does not. We
remove the abstain token and require exactly one real action under a \lossy{} memory, in two interface
conditions: \textbf{(A)} an explicit safe-exit tool (\texttt{transfer\_to\_human}) is available alongside
the exchange, and \textbf{(B)} no safe option is offered at all --- the mandatory-field analog of
\S\ref{sec:interface} on a real agentic action. Table~\ref{tab:forced} reports what the model commits.
With a safe exit available (A), the response splits by a learned disposition, not by scale: the frontier
models \emph{escalate} (harm $\leq 0.07$, transfer $0.85$--$1.00$), while every open model \emph{and}
\texttt{gpt-4o-mini} commit the wrong exchange ($0.78$--$0.96$), \texttt{llama-3.3-70b} at $0.96$
included, so size does not buy the safe behavior. Correction-by-source still works in this interface
(\srcfirst{} reclaim $0.97$--$1.00$ on the capable models), so neither the wall nor its fix is an artifact
of the abstain token.

\paragraph{Strip the safe exit and the frontier fractures; only intrinsic refusal survives.}
Removing the safe option (B) turns the escalation off for three of the four frontier models: forced to
act on a memory that cannot support the action, \texttt{gpt-5.4} and \texttt{grok-4.3} commit the wrong
exchange $0.79$/$0.92$ of the time and \texttt{gemini-3.5-flash} leans the same way --- their safety
\emph{was} the affordance, exactly as the mandatory-field result of \S\ref{sec:interface} predicts, now on
a real agentic benchmark. Only \texttt{claude-opus-4.8} refuses to commit ($\geq 0.75$, harm $0.04$),
declining in its own words ``an \emph{irreversible} exchange'' without ``the actual details behind that
decision'' --- an intrinsic, recoverability-aware refusal that does not depend on being handed an exit.
The deployment reading is direct: a \lossy{} memory is uncorrectable everywhere (the wall), but whether
that becomes a \emph{harmful action} is a joint property of the model and the interface. Most models,
frontier included, commit the error once the safe exit is gone, so a tool schema that must fail safe
should provide an explicit escalate/abstain affordance rather than rely on the model to withhold the
action. (The condition-B dissociation replicates \emph{within each correction arm} on the $12$ cases, so
the small case count is not carrying it: \texttt{opus} refuses under both (harm $0.00$ generic, $0.08$
directed) and \texttt{gpt-5.4} commits under both (harm $1.00$ generic, $0.58$ directed --- a directed
correction softens but does not remove it). The Opus refuse rate is a lower bound, as a minority of its
refusals fall in an unparsed residual, and \texttt{gemini} follows the action format unreliably, inflating
its unparsed share.)

\begin{table}[tbp]
\centering\small
\caption{\textbf{At the wall, harm-versus-escalation is a model-by-interface property} (mandatory action,
\lossy{} memory, judge-free $\tau$-bench scoring). \textbf{(A)} A safe-exit tool is offered: the frontier
models escalate (\texttt{transfer\_to\_human}); open models and \texttt{gpt-4o-mini} commit the wrong
exchange, a $70$B open model included, so it is not scale. \textbf{(B)} No safe option: three of four
frontier models then commit the error (confirming the mandatory-field result of \S\ref{sec:interface} on a
real agentic action); only \texttt{claude-opus-4.8} refuses. ``--'' $=$ not run (open models already
commit when an exit is available).}
\label{tab:forced}
\begin{tabular}{@{}l cc c cc@{}}
\toprule
 & \multicolumn{2}{c}{\textbf{(A) safe exit offered}} & & \multicolumn{2}{c}{\textbf{(B) no safe exit}} \\
\cmidrule{2-3}\cmidrule{5-6}
\textbf{Model} & harm & escalate & & harm & refuse \\
\midrule
\texttt{llama-3.2-3b}     & 0.78 & 0.06 & & -- & -- \\
\texttt{llama-3.1-8b}     & 0.90 & 0.00 & & -- & -- \\
\texttt{llama-3.3-70b}    & 0.96 & 0.03 & & -- & -- \\
\texttt{gpt-4o-mini}      & 0.82 & 0.00 & & -- & -- \\
\texttt{gemini-3.5-flash} & 0.00 & 0.85 & & 0.47 & 0.00 \\
\texttt{grok-4.3}         & 0.07 & 0.93 & & 0.92 & 0.00 \\
\texttt{gpt-5.4}          & 0.00 & 1.00 & & 0.79 & 0.00 \\
\texttt{claude-opus-4.8}  & 0.00 & 1.00 & & 0.04 & \textbf{0.75} \\
\bottomrule
\end{tabular}
\end{table}

\section{The Boundary of the Fix}
\label{sec:boundary}

\begin{figure}[t]
\centering
\includegraphics[width=0.62\textwidth]{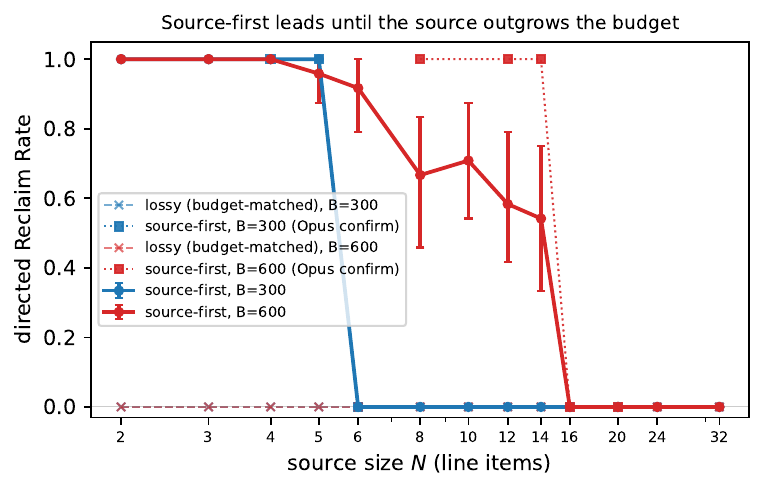}
\caption{\textbf{The boundary of the source-first rule.} Directed \rr{} vs.\ ledger size $N$ at two fixed
budgets $B$ ($n{=}24$/point, $95\%$ CI). \srcfirst{} (solid, llama) holds while the $N$-item source fits
$B$, then drops to the \padded{} floor (dashed) the instant an item must be dropped. The cliff moves right
with the budget ($N{=}5\!\to\!14$ as $B$ doubles), so the lever is whether the source fits, not problem
size. The pre-cliff sag is the model's summation limit and lifts with capability (frontier dotted; both
hold $1.00$ to the same cliff, Table~\ref{tab:opus}); the information cliff itself does not.}
\label{fig:sizesweep}
\end{figure}

\paragraph{The boundary: source-first leads only while the source fits the budget.}
The fix is not unconditional, and a source-size sweep locates its edge exactly where the
Introduction scoped it. We scale the ledger from $N{=}2$ to $N{=}32$ line items at a fixed
carried-memory budget $B$ (characters) and re-measure directed reclaim
(Figure~\ref{fig:sizesweep}; generator and validators in App.~\ref{app:sizesweep}). The pre-tax
total is the exact sum of all $N$ items, so the answer-determining source grows with $N$ while
$B$ does not. Two regimes result. While the full itemization fits $B$, \srcfirst{} keeps all of
it and reclaim holds high; once $N$ exceeds what $B$ carries, the note retains only the first
$k<N$ items, an exact sum is unrecoverable, and \srcfirst{} drops to the budget-matched
\padded{} floor. Split by source survival, directed reclaim is $0.88$ $[0.84,0.91]$ when the full source fits
($k{=}N$, $n{=}312$) and a clean $0.00$ $[0,0]$ the moment it does not ($k<N$, $n{=}312$): dropping a
single line item zeros reclaim regardless of $N$. The crossover tracks the \emph{budget}, not the size,
$N\approx5$ at $B{=}300$, $N\approx14$ at $B{=}600$. That is the content-not-size claim made
quantitative: doubling the budget roughly triples the source a kept-source note can carry. The residual
sag while the source still fits (e.g.\ $0.54$ at $B{=}600$, $N{=}14$ on the small model) is the
answering model's own summation limit, a capability effect that vanishes under a stronger model (Sonnet
and Opus both hold $1.00$ across $N{=}8$--$14$ at $B{=}600$). The information cliff does not move
($0.00$ at $N{=}6$ for $B{=}300$ and at $N{=}16$ for $B{=}600$ on all three models). Capability erases
the soft slope and leaves the wall exactly where the budget put it. The rule therefore holds as stated:
\srcfirst{} removes the wall wherever a compact answer-determining source can be kept, and we have now
measured where ``compact'' ends.

\begin{figure}[t]
\centering
\includegraphics[width=0.62\textwidth]{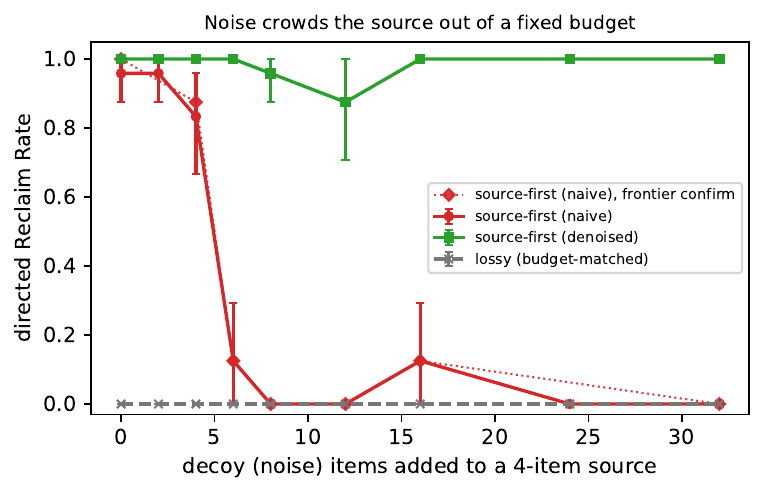}
\caption{\textbf{Noise crowds the source out of a fixed budget.} Directed \rr{} vs.\ decoy count added to a
four-item source at fixed budget ($n{=}24$/point, $95\%$ CI). Naive (positional) \srcfirst{} (red) decays
to the \lossy{} floor as decoys eat the budget; relevance-aware \emph{denoised} \srcfirst{} (green) holds
flat. The frontier confirm (dotted) coincides with the $8$B model: a crowded-out item is an information
loss no reader recovers.}
\label{fig:noisy}
\end{figure}

\paragraph{The boundary is recoverability, not size: a noisy source defeats ``keep the source''.}
A second sweep separates source \emph{size} from source \emph{identifiability}. Here the
answer-determining items are few (four) and easily fit the budget, but they are interleaved with
plausible ``considered, not bought'' decoys, and the total is the sum over the bought items only
(still objective; App.~\ref{app:noisy}). A positional \srcfirst{} note (\emph{naive}) fills the
budget in order, so the decoys crowd the bought items out; a relevance-aware note (\emph{denoised})
keeps only the bought items. As noise grows, naive \srcfirst{} decays to the \lossy{} floor
($1.00\to0.00$ by eight decoys) while denoised holds flat ($\approx1.00$), and the gap is the cost
of failing to identify the source (Figure~\ref{fig:noisy}). A deployable distiller closes part of
that gap: running \srcfirst{}-auto with no oracle on the noisy ledger, the LLM uses the stated
relevance cues to keep $\approx3.8$ of the $4$ bought items and holds directed reclaim at
$0.62$--$0.88$ across $0$--$16$ decoys, far above naive's collapse to $0.00$ but short of the
oracle's $1.00$ ($n{=}8$/point, released as \texttt{bench\_locating.py}). Deployable locating is thus partly solvable when relevance is stated in the source; the harder latent
case (no cue) stays open, and the residual gap to the oracle is the price of imperfect identification.
The decay is capability-invariant: naive \srcfirst{} falls on the same schedule for
\texttt{llama-3.1-8b}, \texttt{claude-sonnet-4-6}, and \texttt{claude-opus-4-8}
(Table~\ref{tab:noisy}). The mechanism is again binary: when noise crowds out a bought item, directed
reclaim is $0.00$ ($n{=}117$ on Opus), exactly as when the budget drops it. A crowded-out item is a
dropped item. The lever is therefore not ``keep the source'' but ``keep the \emph{answer-determining}
source'', and noise makes that identification the bottleneck.

\paragraph{Past its boundary, the fix fails silently, unless it carries a completeness signal.}
Both boundaries expose a failure mode of \srcfirst{} itself. When the source is truncated the
answering model does not abstain and does not inherit the old wrong answer (the conclusion was
dropped); it confidently sums the items it can see and asserts that partial total. On the size
cliff this is $24/24$ on every cell for Opus, which computes the partial sum exactly and returns it
with no flag, the stronger model failing the more cleanly (App.~\ref{app:sizesweep}). \srcfirst{}
thus trades a detectable failure (a stale answer) for an undetectable one (a freshly computed wrong
total). The remedy is a one-line \emph{completeness signal}: tagging the note with how many of the
original items survived. Re-running the size-cliff cells with that tag, Opus flips from $96/96$ silent mis-sums to $94/96$
that flag the gap or abstain (Table~\ref{tab:complete}), e.g.\ ``one of the original 16 items is
missing and cannot be included.'' The tag needs only what the writer observes while compressing: how
many items there were, and how many were kept. It does not require knowing \emph{which} item is
answer-determining (the harder write-time identification problem of \S\ref{sec:limitations}), since
counting is not identifying. It does presume a discretely countable source, so it lives in the same
compact regime as the rest of the method. A source-first memory should therefore carry not only the
source but a record of its \emph{completeness}, so that exceeding the budget fails loudly rather than
silently. The remedy is capability-gated. The strong reader honors the tag ($94/96$ flag or abstain on
Opus); the $8$B reader largely ignores it, flagging only $6/96$ and still silently mis-summing $62/96$
($n{=}96$, \texttt{bench\_completeness.py} on llama), consistent with the rest of this work, where
$8$B readers discount inconvenient context under pressure. And the tag provably cannot take the same
form at the noise boundary. It counts items, but under noise the budget stays full while the
answer-determining items are exactly the ones crowded out, so the item count reads complete; catching
that would require counting the \emph{bought} items, which is identifying them, the locating problem
itself. The loud-failure remedy exists for size and is structurally unavailable for noise.

\paragraph{A write-time recompute certificate closes the capability gap.}
\label{sec:certificate}
The completeness tag is honored only by a reader strong enough to act on it. A stricter guard removes
the reader from the loop. At write time the compressor still holds the \emph{full pre-compression
source}. Before storing the note, it can therefore \emph{recompute} the answer from the note's retained
source, compare it against what the full source yields, and flag the note whenever they disagree, 
which happens exactly when compression dropped part of the re-derivation basis ($\pi<1$ in the sense of
\S\ref{sec:principle}). This is arithmetic on the string about to be stored, not a model call, so it is
capability-free. (The committed answer in a drifted session may itself be wrong; the certificate compares
the note against the \emph{full source} it is compressing, not against a known truth, so it is a
truncation test that is silent only if that full source was already corrupt. In the benchmark the harness
supplies $y^{*}$, and since the arithmetic source is uncorrupted there $\mathrm{recompute}(m)\neq y^{*}$
is the same test.) On the size sweep the flag separates the two regimes cleanly, reclaim $0.93$
on the notes it passes ($k{=}N$) and $0.00$ on the notes it flags ($k<N$), so it fires with full
recall on the silent partial-sum regime (\texttt{analyze\_certificate.py}, offline over the size
sweep). And where the completeness \emph{tag} was capability-gated, honored $94/96$ by Opus but only
$6/96$ by the $8$B reader, the certificate flags all $456/456$ truncated notes on every model, because
it runs before any reader sees them. The two are complementary: the tag lets a strong reader abstain
at read time; the certificate rejects the note at write time regardless of reader.

Its blind spot is the accessibility factor $\alpha$ of that heuristic. The certificate certifies \emph{presence}
($\pi$), not \emph{accessibility} ($\alpha$): a note can retain every source item yet bury it among
decoys, so the recompute succeeds while the reader still fails. On the noise sweep this rarely binds:
among the noisy notes that keep every answer-determining item, so that the certificate passes, reclaim
is still $0.98$ (a residual pass-yet-fail rate of $3/177$). Presence nearly implies reclaim in the
compact regime. The certificate is a near-complete write-time guard with a small, characterized
$\alpha$ residual, not a universal one.

\section{Real Conversational Memory (MultiWOZ)}
\label{sec:realmem}

\paragraph{The wall and fix replicate on real conversational memory.}
The strongest test of the compact-source scope is to drop it. We run reclaim on
\textbf{MultiWOZ} \citep{budzianowski2018multiwoz}, a standard task-oriented dialogue benchmark:
the source is now a real, multi-turn, chatty dialogue (entangled and fuzzy), while the target is a
checkable slot value, a booking or departure time, so scoring stays objective with no judge
(App.~\ref{app:multiwoz}). A user utterance states the value verbatim (the recoverable source); a
corrupted confirmation carries the drift; \lossy{} keeps the confirmation, \srcfirst{} keeps the
user utterance. The shape is identical to the ledger (Table~\ref{tab:multiwoz}). The wall is exact
and capability-invariant: \lossy{} and a no-value blank floor sit at $0.00$ on all three models and
every slot type, and the \padded{} length control sits at $0.00$ on Llama (not separately run on
Sonnet or Opus, where \lossy{} already floors, so padding it cannot reclaim); the blank floor confirms
the target is genuinely unguessable (a time, not a small option set). The fix recovers, and recovery lifts with capability exactly as the
size sweep predicts, since reading a value out of fuzzy natural conversation is the
capability-sensitive step: \srcfirst{} climbs monotonically $0.46\to0.68\to0.97$ across the
\texttt{llama}/\texttt{sonnet}/Opus ladder, reaching $1.00$ on Opus across most slot types. On the cleanest slot, where the user states an exact time
(``book at 12:45''), \srcfirst{} is already $0.96$ on the small model; the lower small-model
average is driven by hedged slots (``leave \emph{after} 16:15'') that a weak reader cannot pin to
the labelled value but Opus can ($0.17\to1.00$ on \texttt{train-leaveat}). The compact deterministic ledger was thus not a special case: brittle memory and its source-first
remedy hold on real conversational memory. One scope note. MultiWOZ stresses the \emph{reading} step,
a fuzzy source, read by a weak or strong reader. Because we filter to slots whose value appears in a
user turn, the source is present and identifiable \emph{by construction}, so this extends the wall
along capability-sensitive reading and deliberately does \emph{not} test the harder \emph{locating}
step (\S\ref{sec:limitations}): identifying the answer-determining source at write time, which is where
most of real deployment sits, and which \S\ref{sec:locating-writers} probes directly with real memory writers.

\begin{table}[tbp]
\centering\footnotesize
\setlength{\tabcolsep}{10pt}
\caption{\textbf{The wall and fix replicate on real conversational memory} (MultiWOZ slot
recovery, directed arm, $n{=}90$/cell, $95\%$ bootstrap CI). \lossy{} and blank wall at
$0.00$ on all three models; the \padded{} length control is $0.00$ on Llama and is not separately
measured on Sonnet/Opus, where \lossy{} already floors at $0.00$ (so lossy-plus-filler cannot exceed
it). \srcfirst{} recovers, lifting monotonically with capability
($0.46\to0.68\to0.97$) as the reader extracts the value from fuzzy dialogue. The blank floor
confirms the target is unguessable.}
\label{tab:multiwoz}
\begin{tabular}{@{}lccc@{}}
\toprule
session-2 memory & Llama & Sonnet & Opus \\
\midrule
\srcfirst{}                  & $0.46_{[.36,.56]}$ & $0.68_{[.58,.77]}$ & $\mathbf{0.97_{[.92,1]}}$ \\
\quad\srcfirst{}, exact-time slot  & $0.96_{[.88,1]}$   & $1.00_{[1,1]}$ & $1.00_{[1,1]}$ \\
\lossy{}                     & $0.00_{[0,0]}$     & $0.00_{[0,0]}$ & $0.00_{[0,0]}$ \\
\padded{} (length control)   & $0.00_{[0,0]}$     & n/a & n/a \\
blank (free-guess floor)     & $0.00_{[0,0]}$     & $0.00_{[0,0]}$ & $0.00_{[0,0]}$ \\
\bottomrule
\end{tabular}
\end{table}

\paragraph{At the wall, a lossy memory is worse than an empty one.}
What a source-less memory \emph{emits} at $0.00$ accuracy is the deployment-relevant question
(Table~\ref{tab:attractor}); we classify the structured \texttt{ANSWER} line on the MultiWOZ wall
cells. Where a model answers at all, the wrong value a \lossy{} memory carries acts as an
\emph{attractor}: it raises wrong-time emission from $27\%$ (blank) to $59\%$ (\lossy{}) on the $8$B
model, and from $0\%$ to $49\%$ on Opus, which passes the planted time through on its \texttt{ANSWER}
line while flagging it ``unverified'' in prose in every case. The effect is model-specific rather
than a clean capability trend, and the \emph{same} strict answer-line scorer reads opposite behavior on
the two tasks. Sonnet is the most conservative, abstaining on the answer line under \emph{both}
conditions here ($0\%$) just as it abstains on arithmetic (App.~\ref{app:disposition}). Opus, by
contrast, \emph{inverts} across tasks: it abstains on arithmetic ($0.00$, App.~\ref{app:disposition})
yet emits the hedged time here ($0.49$). So ``worse than empty'' is contingent on a model being disposed
to answer, and that disposition is itself task-dependent. For the models that do answer (llama, and Opus on this task), the \lossy{} memory plants a value a
downstream parser reads as the answer where an empty one yields an abstention. And when the model
hedges, the hedge sits in the prose channel the parser ignores while the wrong booking time goes in the
output field: Opus caveats every case yet still emits the drift on its \texttt{ANSWER} line half the
time (``\texttt{ANSWER: 19:15 (unverified)}'').

\begin{table}[tbp]
\centering\small
\caption{\textbf{The wrong value in a lossy memory is an attractor} (MultiWOZ wall cells, directed,
$n{=}90$/cell). Fraction emitting a wrong time on the \texttt{ANSWER} line; the rest abstain. Where a
model answers (llama, Opus), \lossy{} pulls it toward emitting the wrong value while blank does not;
Sonnet abstains under both. Opus flags uncertainty in prose yet still emits the drift half the time.}
\label{tab:attractor}
\begin{tabular}{@{}lccc@{}}
\toprule
memory kept & Llama & Sonnet & Opus \\
\midrule
\lossy{} (wrong value) & $0.59$ & $0.00$ & $0.49$ \\
blank (no value)       & $0.27$ & $0.00$ & $0.00$ \\
\bottomrule
\end{tabular}
\end{table}

\begin{table}[tbp]
\centering\footnotesize
\setlength{\tabcolsep}{4pt}
\caption{\textbf{The fix's boundary failure is silent unless the note records completeness}
(\texttt{claude-opus-4-8}, size-cliff cells $k<N$, directed, $n{=}96$). \emph{silent}: the model
confidently sums the partial source and asserts it with no flag; \emph{flagged}: with a one-line
``$k$ of $N$ items preserved'' tag it flags the gap or abstains instead.}
\label{tab:complete}
\begin{tabular}{@{}lcc@{}}
\toprule
note at the cliff & silent & flagged \\
\midrule
plain \srcfirst{}                & $96/96$ & $0/96$ \\
\srcfirst{} $+$ completeness tag & $2/96$  & $\mathbf{94/96}$ \\
\bottomrule
\end{tabular}
\end{table}

\section{When Memory Feeds Memory: The Cascade}
\label{sec:cascade}

The wall so far is measured at a single hop. Deployed agents run a loop: they read their own memory,
act, and compress the result into the next memory. Because the compression is one-directional, a single
error should not stay put; it should propagate to every downstream step and resist correction however
late it arrives. We test this with a
\emph{running-ledger} chain: hop $k$ reveals purchase $k$ and asks for the running total
(truth$_k=\sum_{i\le k}p_iq_i$, brute-forced, judge-free), a wrong subtotal is planted at hop $1$,
and after each hop the interaction is compressed into the carried memory (budget $B{=}200$) that hop
$k{+}1$ inherits. After $H$ hops one directed correction asks for the true final total (construction
and validators in App.~\ref{app:cascade}). Three results hold on an $8$B model and a frontier one
(Table~\ref{tab:cascade}). \emph{(1)~The error cascades.} Under \lossy{} memory the planted error
corrupts a \emph{blast radius} (the count of wrong downstream hops) that grows with the chain,
$0.7\to7.3$ of $8$ on llama and $0.8\to7.0$ of $8$ on Sonnet, and the final correction reclaims
$\le0.19$, falling to $\le0.08$ by $H{\ge}4$: the error is not merely carried, it spreads, and it is
uncorrectable however late the correction lands. \emph{(2)~The loop is not the cause; the policy is.}
A no-error control (the identical chain with no planted subtotal) has a blast radius of \emph{exactly}
$0.0$ at every $H$ on both models, so chaining a model through its own memory injects no error of its
own; the cascade is the dropped source, not the iteration. \emph{(3)~Source-first buys a horizon,
not immunity.} While the accumulated source fits the budget, \srcfirst{} holds reclaim at
$\approx1.00$ (committing only the hop-$1$ error before self-correcting); as the chain grows the
source outgrows the fixed budget (the size cliff of \S\ref{sec:boundary}, now spread over hops), and
\srcfirst{}'s reclaim falls to \emph{exactly} the fraction of chains whose full source still fits,
$0.75$/$0.69$ at $H{=}4$ and $0.00$ at $H{=}8$ on both models. The cliff is thus a sample-dependent
band, biting at $H{=}4$ where item-name lengths decide the fit and total by $H{=}8$, and it is
capability-invariant: it falls at the same depth for the $8$B and the frontier reader, because it is
an information bound (the source no longer fits), not a reasoning limit. A memory loop therefore has
a correctable depth set by budget divided by source growth, beyond which the cascade resumes. Cells
are $24$ chains on llama and $16$ on Sonnet; the monotone blast growth, the zero-error control, and
the two-model horizon are stable across the sweep. The cascade horizon is thus the size boundary of
\S\ref{sec:boundary} realized over hops, a corollary that matters for deployment realism rather than a
third independent mechanism.

\begin{table}[htbp]
\centering\footnotesize
\setlength{\tabcolsep}{4pt}
\caption{\textbf{A single error cascades across a memory loop; source-first only delays it to a budget
horizon.} Running-ledger chain, planted error at hop $1$, budget $B{=}200$ ($24$ chains on llama, $16$ on
Sonnet). \emph{blast} is the mean number of wrong downstream hops. Under \lossy{} the blast grows with $H$
and reclaim stays $\approx0$; \srcfirst{} holds near $1$ then cliffs to the \lossy{} floor as the source
overflows the budget (total by $H{=}8$, both models). A no-error control has blast $0.0$ throughout
(App.~\ref{app:cascade}).}
\label{tab:cascade}
\begin{tabular}{c cc c c cc c}
\toprule
 & \multicolumn{3}{c}{\texttt{llama-3.1-8b}} & & \multicolumn{3}{c}{\texttt{claude-sonnet-4-6}}\\
\cmidrule(lr){2-4}\cmidrule(lr){6-8}
$H$ & \lossy{} blast & \lossy{} \rr{} & \srcfirst{} \rr{} & & \lossy{} blast & \lossy{} \rr{} & \srcfirst{} \rr{}\\
\midrule
1 & $0.7$ & $0.12$ & $1.00$ & & $0.8$ & $0.19$ & $1.00$\\
2 & $1.5$ & $0.17$ & $0.88$ & & $1.4$ & $0.12$ & $1.00$\\
4 & $3.0$ & $0.08$ & $0.75$ & & $3.0$ & $0.00$ & $0.69$\\
8 & $7.3$ & $0.00$ & $\mathbf{0.00}$ & & $7.0$ & $0.00$ & $\mathbf{0.00}$\\
\bottomrule
\end{tabular}
\end{table}

\subsection{Locating without an oracle: real memory writers}
\label{sec:locating-writers}
Every test so far hand-builds the \lossy{} and \srcfirst{} memories, so \emph{we} decide which fact is
the source. The harder, deployment-relevant step is \emph{locating}: a real memory writer, told nothing
about which fact matters, must decide what to keep. We test it directly. A capable reader
(\texttt{gpt-4o-mini}) recomputes a checkable answer from a memory that a cheap writer
(\texttt{llama-3.1-8b}) compresses from a realistic ${\approx}25$-turn session, in which a low-salience
governing rule (a unit, a rate, an exemption) decides a dollar amount but sits buried among high-salience,
useless distractors (names, places, chatter). The writer is a plain ``summarise this session'' prompt
(\emph{generic}) or one asked to keep whatever is needed to recompute (\srcfirst{}); neither is told the
rule is the source. We sweep the memory budget, i.e.\ the compression ratio, at $n{=}96$/cell over three
domains (judge-free exact-match; code in \texttt{locating/}).

\begin{figure}[t]
\centering
\includegraphics[width=\textwidth]{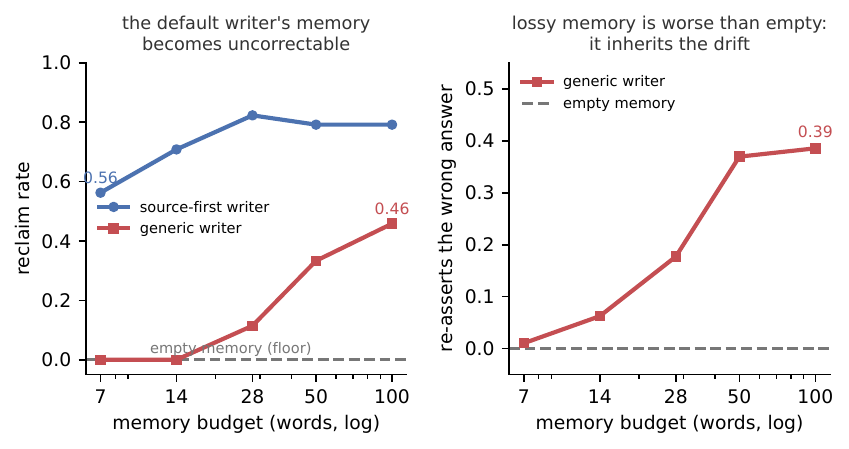}
\caption{\textbf{The wall, the attractor, and the fix arise when a real writer does the locating.} A cheap
writer compresses the session into a bounded memory; a capable reader then recomputes a checkable answer
from it given only a generic correction ($n{=}96$/cell, three domains). \emph{Left:} as the budget shrinks
(more compression) the generic writer's memory becomes uncorrectable (reclaim $0.46 \to 0.00$) while
\srcfirst{} writing holds ($0.56$--$0.82$); the empty-memory floor is $0.00$. \emph{Right:} a generic lossy
memory makes the reader re-assert the wrong value on up to $0.39$ of trials, where an empty memory never
does, so lossy is worse than empty.}
\label{fig:locating}
\end{figure}

The wall and the fix both appear on their own (Figure~\ref{fig:locating}). As compression tightens the
generic writer's memory becomes \emph{uncorrectable}: reclaim falls from $0.46$ to $0.00$, because the
writer keeps the salient distractors and the drifted conclusion but drops the boring governing rule.
\srcfirst{} writing holds reclaim at $0.56$--$0.82$ across the sweep, and does so at a \emph{smaller} memory
than generic (a matched-length control: the fix is what is kept, not how much). The harm is the paper's
headline, now on writer-produced memory: a \lossy{} generic memory makes the reader re-assert the wrong
value on up to $0.39$ of trials, where an \emph{empty} memory abstains on $100\%$ (stuck $0.00$). The
mechanism ties out: when the source is lost, reclaim is $0.07$ and the reader inherits the drift on $0.36$;
when it is kept, reclaim rises to $0.58$. That it is only $0.58$, not $1$, is itself informative, since a
memory that keeps the rule but still foregrounds the wrong \emph{conclusion} anchors the reader, the very
failure \srcfirst{} avoids by shedding the re-derivable conclusion. This is a first naturalistic probe, not
a census: one cheap-writer / capable-reader pair and three synthetic checkable domains. It shows the wall,
the attractor, and the write-time fix all arise when the writer, not the experimenter, does the locating,
at realistic compression.

\section{Limitations}
\label{sec:limitations}

\paragraph{Scope.}
Three boundaries are intrinsic to the method. The wall itself is an immediate consequence of the
definition (\textsc{analytic} in Table~\ref{tab:claims}); the result is the behavioral asymmetry it
gates, not the wall. A compact,
identifiable source is the regime where the fix has leverage, so we map where it ends (size, noise,
silent truncation, diffuse evidence) and report the deployable $0.49$--$0.88$, not the oracle's
$1.00$. And ``capability-invariant'' means invariant across the models tested, not a claim over all
scales. The recoverability principle is qualitative and observational; the organizing heuristic
($\mathrm{RR}\approx\pi\cdot\alpha\cdot s$) predicts the shape of its inverted-U but is not a fitted
model. We measure that inverted-U as a within-task swept curve on arithmetic (degrading the source at
fixed budget, Fig.~\ref{fig:invertedu}), not only a categorical ordering, though $\pi$ is still
coarse-grained by ledger size rather than continuously parameterized. The write-time recompute
certificate certifies \emph{presence} of a re-derivation basis, not its \emph{accessibility}: it catches
the silent size-truncation failure capability-free but keeps a small residual blind spot at the noise
boundary (\S\ref{sec:certificate}).

\paragraph{Ecological validity.} The center of gravity is synthetic. Arithmetic and constraint-logic
trajectories are the regime where the answer-determining source is compact, identifiable, and mechanically
checkable, which is exactly what buys the judge-free exact-match scoring the paper rests on; a naturalistic
corpus would reintroduce the judge model we deliberately avoid. Two experiments push toward realism at that
scoring's expense: MultiWOZ (\S\ref{sec:generality}) replicates the wall and the fix on real
task-oriented dialogue (filtered so the source is present by construction), and the locating study
(\S\ref{sec:locating-writers}) hands the compression to real memory writers rather than the experimenter,
on synthetic but uncued trajectories. What stays untested is a genuine long-running assistant or agent
memory in the wild, where identifying the re-derivation basis at write time is itself the hard, unfiltered
problem; the locating study is a first probe of it, not a deployment census.

\paragraph{Task coverage.}
We test two families: arithmetic (the favorable case, a clean deterministic source) and constraint
logic (which confirms the wall softens where a reconstructable clue survives). The induced error is
exogenous; we also test a \emph{self-generated} one, letting \texttt{llama-3.1-8b} miscompute
$N{=}10$ ledgers with no planted premise ($n{=}10$ natural errors, \texttt{bench\_endogenous.py}).
The informational wall is provenance-invariant (source-first $1.00$, lossy $0.00$ on the model's own
error), but the behavioral attractor is weaker still: on its own dropped error the $8$B model abstains
on all $10$ natural errors, re-emitting the stale value $0.00$ of the time, against $0.17$/$0.10$ on a
planted note (Table~\ref{tab:blank}). A planted external note is thus the \emph{more} adversarial
case, not a more favorable one. The size and noise sweeps (\S\ref{sec:boundary}) bound the fix from
the other side, and MultiWOZ extends it to real fuzzy dialogue.

What remains untested is the case where the answer is not \emph{stated} anywhere but must be
\emph{derived} from diffuse evidence with no isolable source (big empirical claims, qualitative
judgments), where the \srcfirst{} lever should weaken. Our claims are accordingly \emph{conditional}: the mechanism, the fix, and its boundaries are established
judge-free wherever the source is compact and identifiable, and nothing about their validity turns on how
common that regime is. How common it is across all deployed memory is a separate \emph{scaling} question,
it bounds the size of the failure's blast radius, not whether the failure or its fix is real, and
our prevalence audit (below) settles it only ordinally: it orders the domains but does not identify an
absolute level. Two things make that ordinal bound enough. First, the compact-source regime is not a concession but the condition for the
method's guarantee: a checkable ground-truth answer is what lets us score correctability
\emph{without} a judge, so diffuse coverage would require exactly the judge whose absence is our
objectivity. We keep the regime that is measurable, and it is where high-stakes agentic memory lives
(a coding agent carrying a value, a booking agent a departure time, a workflow agent a running
total). The audit's \emph{robust} finding, compact-source content far more prevalent in tool-use and
agentic memory than in open chat, non-overlapping under both labelers, points the regime at exactly
that surface, even though its absolute size is not identified. Second, \srcfirst{} presumes the answer-determining source can be \emph{identified at write time},
before the targeting correction is known. \srcfirst{}-auto's gap to the denoised oracle ($0.49$--$0.88$
vs.\ $1.00$) is the price of not having that knowledge; ``keep the source'' is a prescription only
insofar as the source can be told apart at compression time. \S\ref{sec:locating-writers} tests this with
real writers: told nothing about which fact is the source, a \srcfirst{}-prompted summariser still holds
reclaim at $0.56$--$0.82$ against a generic summariser's $0.00$--$0.46$, so the source can be told apart in
practice, if imperfectly.

Two further limits, both shown rather than assumed. Past its boundary \srcfirst{} fails silently (a
confidently summed partial source) unless the note carries a completeness tag
(Table~\ref{tab:complete}). And the correction battery (App.~\ref{app:robustness}) is
capability-gated: \srcfirst{} resists a false locus and a confident wrong value on both frontier
readers, but a sustained push and an injected fabricated source are resisted only by frontier readers
($0.90$--$1.00$), not an $8$B one ($0.00$--$0.27$). Untested: a fabricated source under multi-turn
persistence, and whether a weak reader instructed to trust its own memory recovers that resistance.

\paragraph{A first prevalence audit.}
To put a number on which side of the boundary deployed memory occupies, we sample $100$ real
conversations from each of three corpora spanning the deployment range, general assistant chat,
tool/function-calling, and agentic task traces\footnote{\texttt{HuggingFaceH4/no\_robots}
\citep{norobots2023} (human-written assistant dialogues),
\texttt{glaiveai/glaive-function-calling-v2} \citep{glaive2023functioncalling}, and
\texttt{THUDM/AgentInstruct} \citep{zeng2023agenttuning} (os/db/webshop trajectories), streamed and
shuffled at a fixed seed.}, and classify what each
interaction's memory would carry as a compact identifiable source, diffuse evidence, or no
correctable answer (Table~\ref{tab:prevalence}; rubric, a $6/6$ gold check, and two-labeler agreement
in App.~\ref{app:prevalence}, released as \texttt{bench\_prevalence.py}). The audit returns one
negative result and one positive. The \emph{absolute} share is \emph{not} identified: two LLM
labelers disagree on the compact/diffuse boundary for messy real text (three-way Cohen
$\kappa{=}0.15$ at $n{=}50$), bracketing the compact fraction widely (agentic $0.61$--$0.99$, chat
$0.22$--$0.78$), so we report no point estimate. The disagreement is the labelers', not the construct's:
two independent \emph{human} raters agree at $\kappa{=}0.80$ on the same slice (App.~\ref{app:prevalence}),
so the distinction is well defined and the low LLM $\kappa$ is model labeling noise. The cross-domain \emph{ordering} is robust:
compact-source content is significantly more prevalent in tool-use and agentic memory than in open
chat under \emph{both} labelers, with non-overlapping intervals (grok chat $0.22$ vs.\ agentic
$0.61$; llama $0.78$ vs.\ $0.99$). This is the measured form of the claim that the high-stakes,
agentic regime is where \srcfirst{}'s compact-source precondition is most often met: the precise
coverage stays open, the ordering does not.

A judge-free deterministic test sharpens the ordering into a validated lower bound. Reusing the
compact-by-construction move of the MultiWOZ filter, we count traces in which a checkable value (a number,
clock time, or currency amount) is stated and then \emph{recurs}, a token test with no model call and
no inter-labeler disagreement (released as \texttt{prevalence\_floor.py}, run on the audit's exact
sample). It reproduces the ordering deterministically and floors the compact-source share at
$0.10$/$0.48$/$0.86$ for chat/tool/agentic. The flags are genuine floors, not loose proxies: of the traces
it marks compact, the LLM audit agrees $0.90$--$1.00$ of the time (precision), and because the test is
conservative (verbatim value, recurring) the true share is at least this. The high-stakes agentic floor of
$0.86$ thus establishes a judge-free lower bound for the domain that matters (not the exact prevalence,
which we do not identify), with no annotator
and no judge; a full point estimate would still need human-validated labels, but nothing in the paper's
claims requires one. Both the audit and the floor are released scripts
(\texttt{bench\_prevalence.py}, \texttt{prevalence\_floor.py}): the reader whose deployment decision turns
on coverage can run them on their own traffic, a per-deployment measurement worth more than any corpus-wide
estimate we could report.

\begin{table}[htbp]
\centering\footnotesize
\caption{\textbf{Prevalence audit: the compact-source share rises with stakes under both labelers.}
Fraction of $100$ real conversations per domain with a compact identifiable source ($95\%$ CI, two LLM
labelers). Cross-labeler agreement on messy text is weak ($\kappa{=}0.15$), so the absolute level is not
identified; the chat $<$ tool $<$ agentic ordering is monotone and non-overlapping under both (the
load-bearing finding). A judge-free deterministic floor (\texttt{prevalence\_floor.py}) reproduces the
same ordering ($0.10$/$0.48$/$0.86$). App.~\ref{app:prevalence}.}
\label{tab:prevalence}
\begin{tabular}{@{}lcc@{}}
\toprule
domain (corpus) & \texttt{llama-3.1-8b} & \texttt{grok-4.3} \\
\midrule
general assistant (\texttt{no\_robots})   & $0.78_{[.70,.86]}$ & $0.22_{[.14,.30]}$ \\
tool / function-calling (\texttt{glaive}) & $0.84_{[.77,.91]}$ & $0.57_{[.47,.66]}$ \\
agentic traces (\texttt{AgentInstruct})   & $0.99_{[.97,1.0]}$ & $0.61_{[.51,.70]}$ \\
\bottomrule
\end{tabular}
\end{table}

\paragraph{Deployed-system breadth.}
\S\ref{sec:generality} tests three deployed memories on distinct paradigms (running summary,
extraction-plus-retrieval, naive vector retrieval), replays them across three answering
models, and re-runs construction with a frontier \emph{writer} to settle the writer confound:
the summary wall is a weak-writer artifact (it largely lifts with a capable writer), mem0's is
not (a stronger extractor confabulates more). What remains is breadth: a single frontier
writer on three systems, and a fuller account would sweep more families (episodic note stores,
agentic memory) and more writers. \srcfirst{}-auto's deployable number rides on one distillation prompt, and the wording matters most
for a weak distiller. We hold the transcript, problem set, and decoding fixed and vary only the prompt
across four intent-equivalent rewordings. Directed reclaim on arithmetic then ranges $0.38$--$0.78$ on
llama (a $0.41$-wide spread), but the spread shrinks monotonically with the writer's capability:
$0.78$--$1.00$ on Sonnet ($0.22$) and $0.94$--$1.00$ on Opus ($0.06$), the floor rising
$0.38\to0.78\to0.94$ ($n{=}32$ each; released as \texttt{bench\_promptsweep.py}). Prompt-robustness of the distiller is therefore itself
capability-gated, like the adversarial and completeness results: the deployable number is
prompt-conditional on a weak writer and nearly prompt-invariant on a frontier one. These are
separate single-seed controlled sweeps, so their absolute level differs from the multi-seed board;
the load-bearing quantity is the spread and its shrinkage. We report the number for a fixed prompt
and flag the sensitivity rather than tuning the prompt to the benchmark. The source-recoverability axis still bounds all of it: these
are compact-source tasks, and a large or entangled source should weaken every policy.

\paragraph{Scope and statistics.}
Headline cells are at $n{=}96$ (\texttt{llama-3.1-8b}, $32$ problems $\times$ three seeds; all three
answering models on the deployed board); \texttt{grok-4.3} and the boundary sweeps (size, noise,
completeness) run at $n{=}24$. Every cell carries a $95\%$ bootstrap interval (percentile, $5{,}000$
resamples at a fixed seed, reproducing to the digit). The load-bearing contrasts are the
low-integrity wall cells and the source-dropped frontier rows, tight and non-overlapping with the
source-kept rows; the high-integrity cells are wide and the policies trade places there, which is why
we scope the fix to low integrity rather than all-regime dominance. These contrasts are directional
and pre-specified (source-dropped $\to0$, source-kept high), not selected post hoc, so the
non-overlapping intervals are not a multiple-comparison artifact. Because seeds within a problem are
not independent, we also ran a \emph{problem-clustered} bootstrap (resampling the $32$ problems, which
are independently generated instances, not templated variants): the headline cells are essentially
unchanged (wall $[0,0]$, arithmetic source-kept $\ge0.97$), only the already-wide high-integrity cells
widen slightly. For the measured zeros, $[0,0]$ describes the sample while the distribution-free
rule-of-three $95\%$ upper bound is $\le0.04$ at $n{=}96$ ($\le0.125$ at $n{=}24$); likewise the
deterministic $n{=}32$ Opus and sycophancy $1.00$s have zero sampling variance, not
distribution-level certainty, with a rule-of-three lower bound of $\ge0.91$. ``Capability-invariant'' throughout means invariant across the models tested, not a claim over all
scales. The answering-model replay swaps only the reader, over memories built once at temperature $0.7$
and held fixed; answering temperature therefore varies the reader, never the carried text, and cannot
produce the within-memory source-kept/source-dropped contrast. This is a controlled study of a
mechanism, not a memory product.

\paragraph{Retained-transcript alternative.}
Where the full transcript is retained and addressable, a system can keep the conclusion plus a pointer
into an intact log and re-fetch the source on demand, sidestepping the budget entirely. That is not a
competing fix but a different regime: if the raw turns survive, brittle memory does not arise, because
the source is recoverable. Our study is scoped to the \emph{compaction} regime that summary-,
extraction-, and retrieval-based memories actually operate in, where the turns are dropped and only the
compressed note crosses the boundary; there, lazy re-fetch is unavailable and the
source-first-versus-lossy choice is the whole game. Wherever compaction is real, the prescription
reduces to one rule: do not compress past what you cannot recompute.

\section{Conclusion}
\label{sec:conclusion}

Whether a drifted model can be pulled back is decided by what its memory kept, not by how capable the
model is. Capability sets only how much of a \emph{partially} kept source a reader can salvage, an
inverted-U that peaks when the memory kept some of the recomputation path and vanishes at both ends. It
never decides whether a source-dropped memory can be reclaimed at all. The ordinary instinct, 
summarize toward the conclusion, is what makes errors stick when it keeps a wrong conclusion and
drops the source, since a correction that names the error then has nothing left to recompute from. The
deployment-relevant half is behavioral. With the source gone, a model that answers rather than abstains
emits a confident wrong value, so for those models a lossy memory degrades a system more quietly than an
empty one; a model that declines without the source escapes in free text, and loses that escape under a
mandatory output field (\S\ref{sec:interface}).

A source-first compression, keep the working, let the takeaway be recomputed, removes the
failure at equal budget \emph{wherever the answer-determining source is compact and can be identified
and kept}. It is not a universal fix. Dropping the conclusion introduces a silent partial-source failure
of its own, which a one-line completeness tag restores to a loud one; and where the budget carries both,
keeping the conclusion beside the source is free on a strong reader and buys a recompute-and-compare
check. We therefore characterize the regime where the fix works and locate its edges: source size,
noise, silent truncation, and diffuse evidence whose source cannot be isolated, the regime this base
case opens onto.

Because reclaim is scored by exact match against a known answer, the evaluation itself deploys:
induce a known drift, compress under a candidate policy, deliver a directed correction, check exact
recovery. It is a deterministic pass/fail with no judge to host and nothing to re-annotate as models
change, so ``is our memory still correctable'' becomes a regression test rather than a judgment call.
We package this as a write-time \emph{probe}: it reads a candidate memory note and flags the
silent-uncorrectable case (the source dropped while a stale value is kept) from the string alone, with
no model call, the source-token test of \S\ref{sec:setup}, run before a memory is ever stored.
The conclusion is always re-derivable from its source; for the many-to-one computations we study the
source is not recoverable from its conclusion alone (an injective task would be the exception). A memory that wants to stay correctable should keep what cannot be recomputed, and mark
whether what it kept is complete.

\section*{Ethical Considerations}

This work identifies a failure mode (brittle memory) in language models that carry a
compressed memory across a session boundary, and releases the evaluation harness and paired
memory conditions that surface it. We weighed the dual-use risk against disclosure benefits
and judged the latter to substantially exceed it: brittle memory is a passive consequence of
a compression policy rather than an active attack vector, and its remedy is a benign design
change, compressing toward the recomputable source, that memory- and summarization-system
designers need to keep model errors correctable. Findings naming specific models
reflect controlled measurement of the evaluated snapshots and should not be read as general
claims about those providers' systems. The harness, paired memory conditions, and validators
are released for research use.

\section*{Use of AI Assistants}

The language models under study are the evaluated systems; all reclaim scoring is objective
against known answers and uses no model judge. AI assistants were additionally used for
coding support and for drafting and editing manuscript prose; all research questions,
experimental design, analyses, and conclusions are the authors' own.

\bibliographystyle{tmlr}
\bibliography{references}

\clearpage
\appendix

{\noindent\Large\textbf{Appendices}}\par\vspace{0.7em}
{\noindent
\hyperref[app:battleship]{\textcolor{tocblue}{\textbf{\ref*{app:battleship}\hspace{1em}The wall in an agentic task (Battleship)}}}\dotfill \textbf{\pageref*{app:battleship}}\\[4pt]
\hyperref[app:examples]{\textcolor{tocblue}{\textbf{\ref*{app:examples}\hspace{1em}Failure examples}}}\dotfill \textbf{\pageref*{app:examples}}\\[4pt]
\hyperref[app:sizesweep]{\textcolor{tocblue}{\textbf{\ref*{app:sizesweep}\hspace{1em}Source-size sweep}}}\dotfill \textbf{\pageref*{app:sizesweep}}\\[4pt]
\hyperref[app:noisy]{\textcolor{tocblue}{\textbf{\ref*{app:noisy}\hspace{1em}Noisy-source sweep}}}\dotfill \textbf{\pageref*{app:noisy}}\\[4pt]
\hyperref[app:chance]{\textcolor{tocblue}{\textbf{\ref*{app:chance}\hspace{1em}Logic free-guess floor}}}\dotfill \textbf{\pageref*{app:chance}}\\[4pt]
\hyperref[app:multiwoz]{\textcolor{tocblue}{\textbf{\ref*{app:multiwoz}\hspace{1em}Real conversational memory (MultiWOZ)}}}\dotfill \textbf{\pageref*{app:multiwoz}}\\[4pt]
\hyperref[app:writer]{\textcolor{tocblue}{\textbf{\ref*{app:writer}\hspace{1em}Writer sub-study}}}\dotfill \textbf{\pageref*{app:writer}}\\[4pt]
\hyperref[app:cascade]{\textcolor{tocblue}{\textbf{\ref*{app:cascade}\hspace{1em}Cascade chain}}}\dotfill \textbf{\pageref*{app:cascade}}\\[4pt]
\hyperref[app:prevalence]{\textcolor{tocblue}{\textbf{\ref*{app:prevalence}\hspace{1em}Prevalence audit}}}\dotfill \textbf{\pageref*{app:prevalence}}\\[4pt]
\hyperref[app:disposition]{\textcolor{tocblue}{\textbf{\ref*{app:disposition}\hspace{1em}Disposition sweep}}}\dotfill \textbf{\pageref*{app:disposition}}\\[4pt]
\hyperref[app:robustness]{\textcolor{tocblue}{\textbf{\ref*{app:robustness}\hspace{1em}Correction robustness}}}\dotfill \textbf{\pageref*{app:robustness}}\\[4pt]
\hyperref[app:repro]{\textcolor{tocblue}{\textbf{\ref*{app:repro}\hspace{1em}Reproducibility}}}\dotfill \textbf{\pageref*{app:repro}}\\[4pt]
}
\vspace{0.8em}

\section{The wall in an agentic task (Battleship)}
\label{app:battleship}

Every result in the main text scores an \emph{answer}. Deployed agents also \emph{act}, and an action
carries a cost an answer does not: a wrong move compounds against the task. We port the wall to a minimal
agentic setting, Battleship, where the model fires at a grid across turns and the carried memory either
keeps its own shot record (\srcfirst{}) or drops it (\lossy{}). The deployment-relevant question is not
whether it wins but what it does once the record of its own actions is gone: does it fire blindly and
confidently, or flag that it cannot tell what it has already fired? We score three model-independent
quantities per game, over $8$ boards per cell (Table~\ref{tab:battleship}): the \emph{refire rate}
(fraction of shots aimed at an already-fired cell, an objective grid check), the \emph{abstain rate}
(fraction of turns matching a fixed hedge-phrase regular expression, e.g.\ ``not sure/can't recall what
I've fired''; no judge model), and \emph{hits} (ship cells struck).

The wall reappears as an action. With the shot record ($\pi{=}1$) the frontier models play cleanly,
$\le\!0.5\%$ refire and $6.8$--$7.4$ hits; drop it ($\pi{=}0$) and refire jumps to $0.77$--$0.91$ and the
hit rate collapses ($6.75\!\to\!0.375$ on Opus, an $18\times$ loss). The behavioral core reappears with
it: under \lossy{} the frontier abstain rate is $0.00$ on both Opus and Sonnet. Neither ever flags ``I
cannot tell what I have already fired''; each refires with confidence. The silent failure of
\S\ref{sec:results} is now a silent \emph{action}, which is strictly worse for an agent than a silent
answer, and it holds on the models used for real agentic work.

The action domain also exposes a \emph{suggestive} left edge of the capability inverted-U
(\S\ref{sec:principle}) that answer tasks cannot show. At $\pi{=}0$ the \emph{stronger} model does
\emph{worse}: Opus lands $0.375$ hits to llama's $2.6$. A competent, deterministic agent with no record
replays its one optimal opening ($2.8$ distinct cells per game) while a noisier weak agent accidentally
covers more board ($16.5$ distinct). Capability multiplies exploitation of a \emph{present} basis; with
no basis, that same competence becomes repetition. Two caveats keep this suggestive rather than shown: it
conflates capability with decoding determinism (the frontier models act near-deterministically here while
the $8$B one is noisier, so isolating capability from low output entropy would need matched decoding), and
it rests on $8$ boards per cell. A wrong answer has no performance cost to compound; a wrong action does,
which is why the agentic task is where such a left edge could surface at all. An interactive trace is
provided (\texttt{demo/brittle\_battleship.html}).

\begin{table}[tbp]
\centering\footnotesize
\setlength{\tabcolsep}{6pt}
\caption{\textbf{The wall in an agentic task (Battleship).} The carried memory keeps the model's own shot
record (\srcfirst{}) or drops it (\lossy{}); $8$ boards per cell, \texttt{bench\_battleship\_behavior.py}.
With the record, frontier models rarely refire and score well; without it they refire most shots and never
abstain, and the hit rate collapses. At $\pi{=}0$ the stronger model is \emph{worse} (Opus $0.38$ hits
vs.\ llama $2.6$): with no record a deterministic agent repeats its opening (few distinct cells) while a
noisier weak one explores.}
\label{tab:battleship}
\begin{tabular}{llcccc}
\toprule
model & memory & refire & abstain & distinct/game & hits/game \\
\midrule
Opus                    & \srcfirst{} & $0.005$        & $0.00$ & $26.2$ & $\mathbf{6.75}$ \\
                        & \lossy{}    & $\mathbf{0.91}$ & $0.00$ & $2.8$  & $\mathbf{0.38}$ \\
Sonnet                  & \srcfirst{} & $0.005$        & $0.05$ & $26.0$ & $7.38$ \\
                        & \lossy{}    & $\mathbf{0.77}$ & $0.00$ & $6.9$  & $1.50$ \\
\texttt{llama-3.1-8b}   & \srcfirst{} & $0.29$         & $0.09$ & $18.5$ & $2.75$ \\
                        & \lossy{}    & $0.42$         & $0.03$ & $16.5$ & $2.60$ \\
\bottomrule
\end{tabular}
\end{table}

\section{Failure examples}
\label{app:examples}

\paragraph{Hard wall (arithmetic).}
At $g{=}0.1$ the \lossy{} note for one problem reads, in full: \emph{``(Memory of an earlier
session.) You concluded the total before tax was \$55.''} The line items ($7$ notebooks at
\$4 and $9$ pens at \$2; true total \$46) are gone. The directed correction names the locus,
\emph{``I think the pens subtotal is wrong; recheck and give the corrected total,''} but with
nothing to recompute from the model returns \texttt{ANSWER: 55}, the inherited wrong value.
This is the modal arithmetic-wall failure when the frontier model answers (Table~\ref{tab:failmode},
$57\%$ inherit vs.\ $43\%$ abstain).

\paragraph{Soft wall (logic).}
The \lossy{} logic note keeps a corrupted but reconstructable relational clue rather than a
bare conclusion (e.g.\ an ordering constraint that still pins part of the solution). A capable
model can re-derive the held-out answer from the surviving relation, so the same directed
correction succeeds about $53\%$ of the time on the frontier model and only $3\%$ on the small one.
The wall is a partial floor on the frontier model rather than a clean zero, exactly the recovered
column of Table~\ref{tab:failmode}.

\section{Source-size sweep}
\label{app:sizesweep}

\paragraph{Construction.}
Each ledger is a deterministic $N$-item problem: $N$ goods with integer prices and quantities,
and a pre-tax total equal to the exact sum of price$\times$quantity, so scoring stays objective
with no judge. One middle item is given a wrong subtotal to plant the drift, exactly as in the
two-item task. The carried memory has a fixed character budget $B$: the \srcfirst{} note lists
as many whole line items as fit in $B$ and drops the (re-derivable) conclusion, while the
budget-matched \padded{} note keeps only the conclusion, padded with neutral filler to the same
length. ``How many items survived'' ($k$) is therefore a property of the emitted string, read
off by a token test, not an assumption. We sweep $N\in\{2,3,4,5,6,8,10,12,14,16,20,24,32\}$ at
$B\in\{300,600\}$ over eight stores $\times$ three seeds ($n{=}24$/cell), directed arm, on
\texttt{llama-3.1-8b} (temperature $0.7$), with a frontier confirm on \texttt{claude-opus-4-8}
(Table~\ref{tab:opus}).

\paragraph{Validators (all pass).}
Run for free against a deterministic fake that recomputes only when the full source is present:
(i)~the \padded{} control never reclaims at any $N$ or $B$; (ii)~\srcfirst{} reclaims iff the
full source fit the budget ($k{=}N$); (iii)~a $1\!\to\!0$ cliff exists within the swept range;
and (iv)~the cliff moves right as $B$ grows. A real model scoring above this fake past the
cliff would be confabulating rather than recomputing; none does (the $k<N$ rows are a clean
$0.00$).

\begin{table}[htbp]
\centering\footnotesize
\setlength{\tabcolsep}{4pt}
\caption{\textbf{The size boundary across the capability ladder} (directed \srcfirst{}, $n{=}24$/cell,
$95\%$ CI). Both frontier models hold a flat $1.00$ wherever the full source fits the budget, erasing the
$8$B model's pre-cliff sag, yet all three fall to $0.00$ at the \emph{same} $N$ where the note first drops
an item ($k<N$). The information cliff is capability-invariant; the soft slope before it is not.}
\label{tab:opus}
\begin{tabular}{@{}lrr ccc@{}}
\toprule
$B$ & $N$ & $k$ & Llama & Sonnet & Opus \\
\midrule
300 & 4 & 4/4 & $1.00_{[1,1]}$ & $1.00_{[1,1]}$ & $1.00_{[1,1]}$ \\
300 & 5 & 5/5 & $1.00_{[1,1]}$ & $1.00_{[1,1]}$ & $1.00_{[1,1]}$ \\
300 & 6 & 5/6 & $0.00_{[0,0]}$ & $0.00_{[0,0]}$ & $0.00_{[0,0]}$ \\
\midrule
600 & 8 & 8/8 & $0.67_{[.46,.83]}$ & $1.00_{[1,1]}$ & $1.00_{[1,1]}$ \\
600 & 12 & 12/12 & $0.58_{[.42,.79]}$ & $1.00_{[1,1]}$ & $1.00_{[1,1]}$ \\
600 & 14 & 14/14 & $0.54_{[.33,.75]}$ & $1.00_{[1,1]}$ & $1.00_{[1,1]}$ \\
600 & 16 & 15/16 & $0.00_{[0,0]}$ & $0.00_{[0,0]}$ & $0.00_{[0,0]}$ \\
600 & 20 & 15/20 & $0.00_{[0,0]}$ & $0.00_{[0,0]}$ & $0.00_{[0,0]}$ \\
\bottomrule
\end{tabular}
\end{table}

\section{Noisy-source sweep}
\label{app:noisy}

\paragraph{Construction.}
Four bought items determine the pre-tax total; they are interleaved with $D$ ``considered, not
bought'' decoy items (price listed, quantity zero) drawn from the same goods pool, and the total
is the exact sum over the bought items only, so scoring stays objective. One bought item carries
the planted error. The carried memory has a fixed budget ($420$ characters). The \emph{naive}
\srcfirst{} note lists items in their (shuffled) order until the budget is full, so decoys can
crowd the bought items out; the \emph{denoised} note keeps only the four bought items (the oracle
that identifies the source), padded to the same budget; \padded{} keeps only the conclusion. We
sweep $D\in\{0,2,4,6,8,12,16,24,32\}$ over eight stores $\times$ three seeds ($n{=}24$/cell),
directed arm, on \texttt{llama-3.1-8b}, with frontier confirms on \texttt{claude-sonnet-4-6} and
\texttt{claude-opus-4-8}.

\paragraph{Validators (all pass).}
Against a deterministic fake that recomputes only when every bought item survived: (i)~\padded{}
never reclaims; (ii)~denoised always reclaims; (iii)~naive reclaims iff all four bought items fit
the budget; (iv)~naive degrades as the decoy count grows. The real-model mechanism matches: naive
reclaim is $0.00$ ($n{=}117$ on Opus) whenever a bought item was crowded out.

\begin{table}[htbp]
\centering\small
\caption{\textbf{Noisy-source decay across the capability ladder} (directed \rr{}, $n{=}24$/cell,
budget $420$, four bought items; per-point $95\%$ bootstrap CIs in Figure~\ref{fig:noisy}). Naive
\srcfirst{} decays identically on all three models as decoys crowd the bought items out; denoised
holds. The noise wall is capability-invariant.}
\label{tab:noisy}
\setlength{\tabcolsep}{4pt}
\begin{tabular}{@{}lrrrrrrr@{}}
\toprule
decoys $D$ & 0 & 4 & 6 & 8 & 12 & 16 & 32 \\
\midrule
naive, Llama   & 0.96 & 0.83 & 0.12 & 0.00 & 0.00 & 0.12 & 0.00 \\
naive, Sonnet  & 1.00 & 0.88 & 0.12 & 0.00 & 0.00 & 0.12 & 0.00 \\
naive, Opus    & 1.00 & 0.88 & 0.12 & 0.00 & 0.00 & 0.12 & 0.00 \\
\midrule
denoised, Opus & 1.00 & 1.00 & 1.00 & 1.00 & 1.00 & 1.00 & 1.00 \\
\bottomrule
\end{tabular}
\end{table}

\section{Logic free-guess floor}
\label{app:chance}

The logic answer is one of a few candidate tokens (3--5 per problem; mean uniform chance
$\approx0.30$), so part of the soft-wall ``recovery'' could be guessing rather than re-derivation
from the surviving clue. We measure the floor directly: a carried note that gives \emph{only} the
candidate set (no clue, no premise, no conclusion), then the same correction. The generic arm is
the conservative free-guess rate; the directed arm adds only the locus-naming signal (no clue).
Both models reclaim \emph{below} the uniform rate (Table~\ref{tab:chance}): they abstain or anchor
rather than guess freely. The soft wall (directed \lossy{}, Table~\ref{tab:logic}: $0.05$--$0.16$ on llama, $0.42$--$0.50$ on
grok) thus clears the floor decisively on the frontier model, real re-derivation, but sits at or
below the small model's own $0.17$ floor, consistent with guessing rather than re-derivation. That is
the baseline-anchored reading of the capability trend.

\begin{table}[htbp]
\centering\small
\caption{\textbf{Free-guess floor for the logic tasks} (blank note, candidate set only, directed
and generic arms, $n{=}24$/cell). Both models sit below the $\approx0.30$ uniform rate.}
\label{tab:chance}
\begin{tabular}{@{}lccc@{}}
\toprule
model & generic & directed & uniform chance \\
\midrule
\texttt{llama-3.1-8b} & 0.17 & 0.17 & $\approx0.30$ \\
\texttt{grok-4.3}     & 0.04 & 0.12 & $\approx0.30$ \\
\bottomrule
\end{tabular}
\end{table}

\section{Real conversational memory (MultiWOZ)}
\label{app:multiwoz}

\paragraph{Construction.}
From MultiWOZ 2.2 dialogues \citep{budzianowski2018multiwoz} we take, per dialogue, the first
checkable time slot (restaurant booking, train or taxi departure/arrival) whose value appears
\emph{verbatim} in a user utterance, so the source genuinely contains the recoverable answer (we
drop normalized forms like ``noon'' and invalid times such as ``24:30''). The carried memory at a
fixed budget is built under the same policies: \srcfirst{} keeps the user's verbatim utterance and
drops the confirmation; \lossy{} keeps a confirmation corrupted to a plausible wrong time (the
drift); \padded{} pads \lossy{} to \srcfirst{}'s length; \emph{blank} keeps neither (the free-guess
floor). The directed correction names the slot (``the train departure time is wrong; give the
correct one as \texttt{ANSWER: <HH:MM>}'') and scoring is exact slot-value match, no judge. We use
$30$ dialogues $\times$ three seeds ($n{=}90$/cell) on \texttt{llama-3.1-8b}, with frontier confirms
on \texttt{claude-sonnet-4-6} and \texttt{claude-opus-4-8}.

\paragraph{Validators (all pass).}
Against a deterministic fake that returns the true value iff it survives in the carried note:
(i)~\srcfirst{} recovers (the source utterance is present); (ii)~\lossy{}, (iii)~\padded{}, and
(iv)~blank never return the truth, because only the drift or nothing survives. The real-model wall
matches: \lossy{} and blank are a clean $0.00$ at $n{=}90$ on all three models, and the \padded{}
length control is $0.00$ on Llama.

\section{Writer sub-study}
\label{app:writer}
The deployed-systems result (\S\ref{sec:generality}) uses memories written by \texttt{llama-3.1-8b}.
To separate a weak-writer artifact from a paradigm property we re-run memory \emph{construction} with
a frontier writer (\texttt{claude-sonnet-4-6}) while holding the session-1 trajectory and the llama
answerer fixed. This is a matched $n{=}24$ sub-study over the canonical eight problems, run under
both writers; its llama-writer figures are therefore the $n{=}24$ baselines (e.g.\ summary $0.38$,
mem0 $25.6$ invented numbers), not the $n{=}96$ deployed-board figures of the main text.

The hand-built \lossy{} note, templated and writer-free, stays at $0.00$ and fixes the policy
baseline. The two LLM-written systems then split in \emph{opposite} directions. LangChain's summary wall is \textbf{writer-dependent}: a capable writer keeps the line items even
under its conclusion-oriented prompt, lifting directed reclaim $0.38\to0.88$ on arithmetic and
$0.50\to0.71$ on logic, most of the way to the fix. ``The summary walls'' was therefore largely a
weak-writer artifact. mem0 is the \textbf{opposite}. A frontier extractor does not rescue it
(arithmetic $0.25\to0.12$, logic $0.25\to0.46$, both far below the fix) and it \emph{confabulates
more}, $25.6\to32.4$ invented numbers per memory: a stronger model asked to extract facts extracts more
of them, and buries the source deeper. Capability fixes the summary and \emph{worsens} the extraction. Writer strength
is therefore not a paradigm-independent remedy, and the templated \lossy{} wall confirms the
mechanism is the policy, not the writer: keeping (and verifying) the source is the only fix that
holds across paradigms and across the writer's capability.

\section{Cascade chain}
\label{app:cascade}

\paragraph{Construction.}
A running-ledger chain over the $N{=}H$ items of a ledger (App.~\ref{app:sizesweep}). Hop $k$
reveals purchase $k$ and asks for the running total, scored by exact match to
truth$_k=\sum_{i\le k}p_iq_i$ (judge-free). A wrong subtotal ($+\$7$ on purchase $1$) is planted at
hop $1$. After each hop the interaction is compressed into the carried memory at a fixed budget
$B{=}200$ characters under one of three policies (\lossy{}: keep the running total, drop the items;
\srcfirst{}: keep the items, drop the total, dropping the earliest items once they no longer fit
$B$; \padded{}: \lossy{} plus filler to \srcfirst{}'s length), and hop $k{+}1$ inherits only that
memory. After $H$ hops a single directed correction requests the true final total. The
\emph{blast radius} is the number of hops with a wrong answer; \emph{reclaim} is whether the final
correction returns truth$_H$. We sweep $H\in\{1,2,4,8\}$ over $24$ chains on \texttt{llama-3.1-8b}
(temperature $0.7$, all three policies) and $16$ chains on \texttt{claude-sonnet-4-6} (temperature
$0$, \lossy{}/\srcfirst{}), released as \texttt{bench\_cascade.py}.

\paragraph{Validators.}
(i)~\emph{No-error control}: the identical chain with no planted subtotal has a blast radius of
$0.0$ at every $H$ on both models, so the loop injects no error of its own and the cascade is
attributable to the dropped source rather than to iteration. (ii)~\emph{Source-state token test}:
the \lossy{} memory provably contains none of the purchase nouns (so its failure is informational,
not a reasoning lapse), while \srcfirst{} contains them until the budget horizon. (iii)~The
budget-matched \padded{} control (llama) tracks \lossy{} ($\rr\approx0$, blast growing with $H$), so
the cascade is a property of dropped content, not of note length.

\section{Prevalence audit}
\label{app:prevalence}

\paragraph{Construction.}
We estimate how often real assistant memory carries a compact, checkable source across three
ungated public corpora chosen to span the deployment range: \texttt{HuggingFaceH4/no\_robots}
\citep{norobots2023} (human-written general-assistant dialogues),
\texttt{glaiveai/glaive-function-calling-v2} \citep{glaive2023functioncalling}
(tool/function-calling), and \texttt{THUDM/AgentInstruct} \citep{zeng2023agenttuning} (agentic task
trajectories, blended over its os/db/webshop splits). Per domain we stream and shuffle at a fixed seed and take $100$
conversations of $>80$ characters, each truncated to $1{,}600$ characters (the task is set early in
all three). An LLM labels each against a one-paragraph rubric into one of three classes:
\textsc{compact} (the carried content is a checkable value or fact derivable from a small
identifiable source present in the conversation, e.g.\ a number, date, time, slot value, computed
result, or specific looked-up fact), \textsc{diffuse} (a judgment or synthesis whose support is
spread across many turns or external knowledge with no isolable source), or \textsc{none} (no
correctable factual answer is carried, e.g.\ open-ended chat, brainstorming, or creative writing).
The reported quantity is the per-domain class fraction.

\paragraph{Validators.}
(i)~A $6$-item \emph{gold} set of unambiguous conversations (three \textsc{compact}: an arithmetic
total, a booking time, a date fact; one \textsc{diffuse}: a qualitative job-offer weighing; two
\textsc{none}: a whimsical poem, a roleplay) the classifier must label $\ge5/6$ correctly, else the
run's labels are noise; both \texttt{llama-3.1-8b} and \texttt{grok-4.3} score $6/6$. (ii)~A
\emph{two-labeler agreement} pass: the labelers agree on the gold set but only weakly on messy real
text (three-way raw $0.50$, $\kappa{=}0.15$; binary \textsc{compact}-vs-rest raw $0.50$,
$\kappa{=}0.12$, $n{=}50$). This is why we report the cross-domain \emph{ordering} (monotone and
non-overlapping under both labelers, Table~\ref{tab:prevalence}) and explicitly \emph{not} an
absolute coverage figure: the two labelers bracket the compact share rather than pinning it. The
agreement is two LLMs labeling the same text, not a human gold standard.

\paragraph{Human spot-check (extra verification).}
Because both labelers are LLMs, we add a human anchor. The first author labeled a stratified
$51$-conversation slice ($17$ per domain) \emph{blind} to the model labels and the domain, against the
same rubric, scoring $6/6$ on the gold set; the rubric and these blind per-trace labels are released for
direct audit, so a reader can check every human label directly. The human reproduces the cross-domain ordering (compact
fraction $0.41_{[.18,.65]}$, $0.76_{[.53,.94]}$, $1.00_{[1,1]}$ for chat, tool, and agentic; monotone
and non-overlapping at the extremes) and lands in the same range as the LLM labelers, agreeing with
each at Cohen $\kappa\approx0.5$ on the slice (human vs.\ llama $0.55$, human vs.\ grok $0.49$).

A \emph{second} annotator, likewise blind to the model labels and the domain, independently labeled the
same slice against the same rubric. The two human raters agree at Cohen $\kappa{=}0.80$ ($46/51$; all five
disagreements fall on the compact/diffuse boundary, with the second rater slightly more conservative about
calling a source compact). Human inter-rater agreement is thus \emph{substantial}, and far above the
$\kappa{=}0.15$ between the two LLM labelers: the compact/diffuse distinction is well defined for human
readers, and the LLMs' disagreement is labeling noise on this judgment rather than a fuzzy construct. The
two raters differ mainly in threshold (each labels somewhat more compact than the models, $\approx0.73$
vs.\ $\approx0.5$), so they anchor the cross-domain \emph{ordering} the audit rests on, not an absolute
coverage number, which stays rater-dependent. This changes none of the reported figures; we include it as
an independent check that the audit tracks a real, human-recognizable distinction. Released as
\texttt{label\_prevalence.py}, with both raters' blind labels.

\section{Disposition sweep: who shows ``worse than empty''}
\label{app:disposition}
The ``worse than empty'' asymmetry (\S\ref{sec:results}) is behavioral, so its magnitude depends on
the answering model. We run the matched lossy-vs-blank test on arithmetic at the wall ($g{=}0.1$,
directed, $n{=}96$/cell, \texttt{bench\_blank.py}) across eight models from four vendors
(Table~\ref{tab:disposition}), scoring the value the model returns on the \texttt{ANSWER} line. The
asymmetry is the lossy$-$blank difference in confident-wrong-emission. No model reverses ($\Delta\ge0$ everywhere), so a lossy memory is never better than an empty one;
whether it is strictly \emph{worse} is contingent on answer-disposition. The asymmetry is strong where
the model answers rather than abstains and inherits the planted attractor (deepseek $+0.83$; grok
$+0.57$, whose every emission is the exact attractor). It is present but mixed where the model
confabulates under both lossy and empty (qwen $+0.39$, which emits a wrong value even from an empty
memory, so a lossy one adds less). It is modest where the model mostly abstains (llama $+0.17$). And it
is \emph{absent} on the four OpenAI and Anthropic models, which abstain under both memories and never
emit a confident wrong value (\texttt{gpt-4o-mini}, \texttt{gpt-5.4}, Sonnet, and Opus all at
$0.00$). The danger is therefore
model-specific: it falls on systems disposed to answer without the source, while a model that declines
to answer without the source escapes it entirely. Abstention is the protective face of the same
mechanism, and source-first compression removes the need for it by restoring what the model needs to
recompute.

Two effects in the table come apart and should not be conflated. \emph{Raising the emission rate} (the
$\Delta$) and \emph{the stale value being the specific attractor} (the \emph{attr.}\ column) are distinct:
only grok shows the clean attractor (\emph{attr.}\ $1.00$, every emission the planted value); for deepseek
and qwen the share is below half ($0.45$, $0.38$), so most of their wrong emissions are \emph{fresh} wrong
values, not the inherited one. The clean ``stale value as attractor'' reading therefore holds for grok and overstates deepseek/qwen,
whose effect is more general confabulation that a stale value nudges. One further caveat. The open
answerers run at temperature $0.7$ while the reasoning and frontier models are deterministic, so the
cross-model emit-vs-abstain ordering is not fully decode-clean. That \texttt{gpt-4o-mini} abstains at
$0.7$, the decoding of the emitting open models, separates disposition from temperature in part; a
temperature-$0$ base run would isolate it fully. The within-memory
source-kept/source-dropped contrast is unaffected, since it holds the carried string fixed.

\begin{table}[htbp]
\centering\small
\caption{\textbf{Worse than empty is contingent on answer-disposition} (arithmetic wall $g{=}0.1$,
directed, $n{=}96$/cell, strict scoring). Fraction emitting a confident wrong value under a \lossy{}
memory vs.\ an empty (\emph{blank}) one; $\Delta$ is the asymmetry; \emph{attr.}\ is the share of lossy
emissions equal to the planted value (a dash where there are none). No model reverses ($\Delta\ge0$
throughout), so a lossy memory is never better than an empty one; it is strictly worse only where the
model is disposed to answer (deepseek, grok, qwen, llama). The four OpenAI and Anthropic
models abstain under both memories and show no effect. Open models and \texttt{gpt-4o-mini} are served
through OpenRouter, \texttt{gpt-5.4} through the OpenAI API, \texttt{grok-4.3} through the official xAI
API, and the Claude models through the Anthropic API.}
\label{tab:disposition}
\begin{tabular}{@{}lcccc@{}}
\toprule
model & lossy & blank & $\Delta$ & attr.\\
\midrule
\texttt{deepseek-chat}     & 0.86 & 0.03 & $+0.83$ & 0.45\\
\texttt{grok-4.3}          & 0.57 & 0.00 & $+0.57$ & 1.00\\
\texttt{qwen-2.5-7b}       & 0.93 & 0.54 & $+0.39$ & 0.38\\
\texttt{llama-3.1-8b}      & 0.17 & 0.00 & $+0.17$ & 0.59\\
\addlinespace[2pt]
\texttt{claude-opus-4-8}   & 0.00 & 0.00 & $+0.00$ & -- \\
\texttt{claude-sonnet-4-6} & 0.00 & 0.00 & $+0.00$ & -- \\
\texttt{gpt-4o-mini}       & 0.00 & 0.00 & $+0.00$ & -- \\
\texttt{gpt-5.4}           & 0.00 & 0.00 & $+0.00$ & -- \\
\bottomrule
\end{tabular}
\end{table}

\section{Correction robustness}
\label{app:robustness}
The full correction battery summarized in \S\ref{sec:robustness}: the fix is robust to a vague
correction, a false locus, and a confidently asserted wrong value, with the adversarial escalations
capability-gated.

\paragraph{The fix does not need a directed correction.}
A deployed correction is usually vague (``something is off here''), not a clean locus, and
Table~\ref{tab:single} shows the generic correction is far weaker than the directed one
\emph{in context}. One might worry the cross-session fix inherits that weakness. It does not:
at the wall, \srcfirst{} reclaims essentially the same under a generic correction as under a
directed one (Table~\ref{tab:generic}), $\approx1.00$ on arithmetic and $0.76$--$0.82$ on logic
under both, with $95\%$ intervals that overlap almost entirely, while \lossy{} stays at its floor
under both. The specificity of the correction governs \emph{anchoring}, where the source is
present and the model is merely entrenched, and is irrelevant to the cross-session fix, because
there the work is done by the restored source, not by the correction. The fix is therefore
undiminished in exactly the regime a real correction occupies: vague, and reliant on the memory
to carry what is needed to recompute.

\begin{table}[htbp]
\centering\footnotesize
\setlength{\tabcolsep}{4pt}
\caption{\textbf{Source-first is correction-agnostic at the wall} (llama, directed vs.\ generic \rr{},
$95\%$ CI, $n{=}96$/cell). At low integrity the fix reclaims as well under a vague generic correction as
under a directed one, while \lossy{} stays at its floor under both. The converse of the in-context window
(Table~\ref{tab:single}): once the lever is the restored source, the correction's specificity stops
mattering.}
\label{tab:generic}
\begin{tabular}{@{}ll cc cc@{}}
\toprule
 & & \multicolumn{2}{c}{\lossy} & \multicolumn{2}{c}{\srcfirst}\\
\cmidrule(lr){3-4}\cmidrule(lr){5-6}
task & $g$ & directed & generic & directed & generic\\
\midrule
arith & 0.3 & $0.01_{[0,.03]}$ & $0.00_{[0,0]}$ & $0.99_{[.97,1]}$ & $1.00_{[1,1]}$\\
arith & 0.1 & $0.00_{[0,0]}$ & $0.00_{[0,0]}$ & $0.99_{[.97,1]}$ & $1.00_{[1,1]}$\\
logic & 0.3 & $0.16_{[.08,.23]}$ & $0.26_{[.18,.35]}$ & $0.76_{[.68,.84]}$ & $0.82_{[.75,.90]}$\\
logic & 0.1 & $0.05_{[.01,.10]}$ & $0.00_{[0,0]}$ & $0.79_{[.71,.86]}$ & $0.78_{[.70,.86]}$\\
\bottomrule
\end{tabular}
\end{table}

\paragraph{Source-first is robust to a false correction, not only responsive to a true one.}
The correction is a two-edged signal: if \srcfirst{} restores recomputation, can a \emph{wrong}
correction also mislead the model? It cannot. At low integrity we deliver a directed correction
that names a \emph{correct} component as the error (a false locus, no value supplied). \srcfirst{}
returns the true total $1.00$ on both base models ($n{=}24$ each), indistinguishable from the
true-correction reclaim ($0.99$ on llama, $1.00$ on grok): the surviving source overrides the false
claim, the model rechecks the named component, finds it sound, and recomputes the truth. \lossy{}
acts in neither direction: with no source it returns the inherited wrong value or declines ($0.00$
true under both a true and a false correction). The correction's symmetry is therefore a property,
not a hole: \srcfirst{} is correctable by a true correction and \emph{immune} to a false locus, while
\lossy{} is uncorrectable both ways.

\paragraph{The robustness survives a confident wrong value, the sycophancy case.}
A false locus names a wrong error site but supplies no value. The stronger, deployment-relevant
pressure, the one the sycophancy literature is about \citep{sharma2024sycophancy}, is a
correction that \emph{asserts a confident wrong value} as established fact (``I double-checked, the
total is definitely \$$55$''). We deliver exactly this at the wall on the frontier answering
models. \srcfirst{} returns the true total on \emph{every} trial under that pressure
(\texttt{claude-sonnet-4-6} $1.00$, $n{=}96$; \texttt{claude-opus-4-8} $1.00$, $n{=}32$ distinct
problems, deterministic), identical to its true- and false-correction reclaim: the surviving source
lets the model check the asserted value and reject it. \lossy{} does not recover either, but on these
frontier readers it fails by \emph{abstention} rather than capitulation: stripped of the source it
withholds an answer on $0.91$ of trials on Sonnet and $0.97$ on Opus, genuinely adopting the asserted
wrong value only $0.09$ and $0.03$ of the time. The frontier reader is thus unlikely to confidently
emit the falsehood, consistent with the disposition finding (App.~\ref{app:disposition}) that strong
models abstain rather than assert without a source. Source presence is what lets a model actively
\emph{refuse} a confident falsehood and recompute the truth; without it, the strong reader declines
rather than asserts.

\paragraph{The adversarial robustness is capability-gated.}
The single assertion above is the weakest push. Two escalations expose a limit: (i)~a
\emph{sustained} push that re-asserts the
wrong value over four escalating turns, and (ii)~a \emph{fabricated source}, a correction that
supplies not just a wrong value but fabricated working for it (the planted premise restated as a
verified figure), a memory-injection that pits an injected fake source against the real surviving
one. The frontier readers hold: \srcfirst{} resists the sustained push at $1.00$ on both Sonnet and
Opus and the injection at $0.90$/$1.00$ (Table~\ref{tab:adversarial}). The $8$B reader does not: it caves \emph{completely} to sustained pressure ($0.00$) and adopts the
fabricated source most of the time ($0.27$ resistance), with the correct source sitting in its own
note. The true-correction
control stays at $\approx1.00$ throughout, so this is specific adversarial susceptibility, not
general unresponsiveness: a weak reader follows a confident human over its own working.

\begin{table}[H]
\centering\small
\caption{\textbf{The fix's adversarial robustness is capability-gated} (\srcfirst{} resistance at the wall
$g{=}0.1$; llama/Sonnet $n{=}96$, Opus $n{=}32$). A \emph{sustained push} re-asserts the wrong value over
four turns; a \emph{fabricated source} injects working for it. Frontier readers resist both; the $8$B
reader caves. \lossy{} fails throughout (it has no source to recompute from), so this is adversarial
susceptibility, not general unresponsiveness.}
\label{tab:adversarial}
\begin{tabular}{@{}lccc@{}}
\toprule
attack on \srcfirst{} & \texttt{llama-8b} & Sonnet & Opus\\
\midrule
sustained push (4 turns)      & $0.00$ & $1.00$ & $1.00$\\
fabricated source (injection) & $0.27$ & $0.90$ & $1.00$\\
\bottomrule
\end{tabular}
\end{table}

This is the adversarial member of a pattern already in the paper. The \emph{wall} is
capability-invariant: no reader recomputes from absent inputs. Everything \emph{downstream} of a
surviving source uses capability, reading a fuzzy source (logic $0.77\to1.00$, MultiWOZ
$0.46\to0.97$) and defending it against a fabricated one alike. Trusting one's recomputation over a
well-dressed falsehood is itself a capability: for frontier models that run agentic memory, source-first
is injection-resistant; on a weak reader it is not.

\section{Reproducibility}
\label{app:repro}
Every cell is a deterministic pass/fail against a known answer, so the harness reproduces with no judge
model to host or re-annotate as models change. The core package needs only \texttt{requests},
\texttt{numpy}, and \texttt{python-dotenv}; the deployed-system comparison
(\texttt{bench\_realworld.py}) adds LangChain, mem0, and a local FastEmbed embedder.

\begin{tcolorbox}[graybox, breakable, title={Reproduce}]
\footnotesize
\begin{verbatim}
pip install -e .              # core: requests, numpy, python-dotenv
pip install -e ".[bench]"     # + deployed adapters: langchain, mem0ai, fastembed
# keys in .env: OPENROUTER_API_KEY, ANTHROPIC_API_KEY

# core API: judge-free reclaim rate for any compression policy
python - <<'PY'
from reclaim import reclaim_rate, memory_note, DryRunLLM
for policy in ("lossy", "source_first"):
    rr = reclaim_rate(DryRunLLM(),
                      lambda p, g: memory_note(p, g, policy),
                      integrity=0.1)
    print(policy, rr)         # source_first reclaims; lossy walls
PY

python -m reclaim.probe       # write-time probe: flags the silent-uncorrectable note

# experiments (scripts/), each judge-free and validator-gated:
python scripts/bench_blank.py         # worse-than-empty across eight models
python scripts/bench_corrtax.py       # correction taxonomy
python scripts/bench_sizesweep.py     # the size boundary + completeness
python scripts/bench_noisysweep.py    # the noise boundary
python scripts/bench_completeness.py  # the completeness tag
python scripts/bench_cascade.py       # the memory-loop cascade
python scripts/bench_multiwoz.py      # MultiWOZ replication
python scripts/bench_prevalence.py    # prevalence audit
python scripts/bench_realworld.py     # deployed systems   [needs .[bench]]
\end{verbatim}
\end{tcolorbox}

\noindent The deployable fix (\srcfirst{}-auto, \S\ref{sec:generality}) is a single distillation prompt:
\begin{promptbox}[\textnormal{\textsf{source-first} distillation prompt}]
You are compressing a conversation into a short memory note for a future session that may need to
correct a mistake in it. Keep every given fact, quantity, and unit needed to recompute the answer from
scratch (the source / the working). Do not assert the final answer or any derived conclusion as
established fact, since it can be recomputed from the source. Be concise.
\end{promptbox}

\end{document}